\documentclass[10pt,twocolumn,letterpaper]{article}

\usepackage{cvpr}              

\usepackage{graphicx}
\usepackage{amsmath}
\usepackage{amssymb}
\usepackage{booktabs}

\usepackage{caption}
\usepackage{subcaption}
\usepackage{multirow}
\usepackage{color,colortbl}
\usepackage{wrapfig}
\usepackage{float}

\newcommand{\diag}{\mathop{\mathrm{diag}}}
\makeatletter
\def\blfootnote{\gdef\@thefnmark{}\@footnotetext}
\makeatother

\definecolor{beaublue}{rgb}{0.74, 0.83, 0.9}

\usepackage[pagebackref,breaklinks,colorlinks]{hyperref}

\usepackage[capitalize]{cleveref}
\crefname{section}{Sec.}{Secs.}
\Crefname{section}{Section}{Sections}
\Crefname{table}{Table}{Tables}
\crefname{table}{Tab.}{Tabs.}

\def\cvprPaperID{818} 
\def\confName{CVPR}
\def\confYear{2022}

\begin{document}

\title{Unsupervised Action Segmentation by Joint Representation Learning and Online Clustering}

\author{Sateesh Kumar$^\dagger$~~~~~~~~~~Sanjay Haresh$^\dagger$~~~~~~~~~~Awais Ahmed~~~~~~~~~~Andrey Konin\\M. Zeeshan Zia~~~~~~~~~~Quoc-Huy Tran\\
\\
Retrocausal, Inc.\\
Seattle, WA\\
\url{www.retrocausal.ai}
}

\maketitle

\begin{abstract}
We present a novel approach for unsupervised activity segmentation which uses video frame clustering as a pretext task and simultaneously performs representation learning and online clustering. This is in contrast with prior works where representation learning and clustering are often performed sequentially. We leverage temporal information in videos by employing temporal optimal transport. In particular, we incorporate a temporal regularization term which preserves the temporal order of the activity into the standard optimal transport module for computing pseudo-label cluster assignments. The temporal optimal transport module enables our approach to learn effective representations for unsupervised activity segmentation. Furthermore, previous methods require storing learned features for the entire dataset before clustering them in an offline manner, whereas our approach processes one mini-batch at a time in an online manner. Extensive evaluations on three public datasets, i.e. 50-Salads, YouTube Instructions, and Breakfast, and our dataset, i.e., Desktop Assembly, show that our approach performs on par with or better than previous methods, despite having significantly less memory constraints. Our code and dataset are available on our research website: \url{https://retrocausal.ai/research/}.
\end{abstract}


\section{Introduction}
\label{sec:introduction}
{\blfootnote{$^{\dagger}$ indicates joint first author.\\ \{sateesh,sanjay,awais,andrey,zeeshan,huy\}@retrocausal.ai.}} 

\begin{figure}[t]
	\centering
		\includegraphics[width=0.9\linewidth, trim = 0mm 95mm 155mm 0mm, clip]{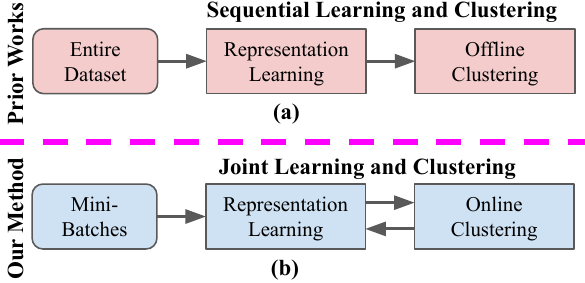}
	\caption{ (a) Previous approaches~\cite{sener2018unsupervised,kukleva2019unsupervised,vidalmata2021joint,li2021action} often perform representation learning and clustering sequentially, while storing embedded features for the entire dataset before clustering them. (b) We unify representation learning and clustering into a single joint framework, which processes one mini-batch at a time. Our method explicitly optimizes for unsupervised activity segmentation and is much more memory efficient.}
	\label{fig:teaser}
\end{figure}

With the advent of deep learning, significant progress has been made in understanding human activities in videos. However, most of the research efforts so far have been invested in action recognition~\cite{tran2015learning,carreira2017quo,wang2018non,tran2018closer}, where the task is to classify simple actions in short videos. Recently, a few approaches have been proposed for dealing with complex activities in long videos, e.g., temporal action localization~\cite{shou2016temporal,shou2017cdc,chao2018rethinking,zeng2019graph}, which aims to detect video segments containing the actions of interest, and anomaly detection~\cite{sultani2018real,gong2019memorizing,haresh2020towards}, whose goal is to localize video frames containing anomalous events in an untrimmed video.

In this paper, we are interested in the problem of temporal activity segmentation, where our goal is to assign each frame of a long video capturing a complex activity to one of the action/sub-activity classes. One popular group of methods~\cite{kuehne2014language,kuehne2016end,lea2016segmental,chen2020action,li2020ms} on this topic require per-frame action labels for fully-supervised training. However, frame-level annotations for all training videos are generally difficult and prohibitively costly to acquire. Weakly-supervised approaches which need weak labels, e.g., the ordered action list or transcript for each video~\cite{huang2016connectionist,richard2016temporal,kuehne2017weakly,richard2017weakly,richard2018neuralnetwork,ding2018weakly,chang2019d3tw,li2019weakly}, have also been proposed. Unfortunately, these weak labels are not always available a priori and can be time consuming to obtain, especially for large datasets.

To avoid the above annotation requirements, unsupervised methods~\cite{malmaud2015s,sener2015unsupervised,alayrac2016unsupervised,sener2018unsupervised,kukleva2019unsupervised,vidalmata2021joint,li2021action} have been introduced recently. Given a collection of unlabeled videos, they \emph{jointly} discover the actions and segment the videos by grouping frames across all videos into clusters, with each cluster corresponding to one of the actions. Previous approaches~\cite{sener2018unsupervised,kukleva2019unsupervised,vidalmata2021joint,li2021action} in unsupervised activity segmentation usually separate the representation learning step from the clustering step in a sequential learning and clustering framework (see Fig.~\ref{fig:teaser}(a)), which prevents the feedback from the clustering step from flowing back to the representation learning step. Also, they need to store computed features for the entire dataset before clustering them in an offline manner, leading to inefficient memory usage.

In this work, we present a joint representation learning and online clustering approach for unsupervised activity segmentation (see Fig.~\ref{fig:teaser}(b)), which uses video frame clustering as a pretext task and hence directly optimizes for unsupervised activity segmentation. We employ temporal optimal transport to leverage temporal information in videos. Specifically, the temporal optimal transport module preserves the temporal order of the activity when computing pseudo-label cluster assignments, yielding effective representations for unsupervised activity segmentation. In addition, our approach processes one mini-batch at a time, thus having substantially lesser memory requirements.

In summary, our contributions include:
\begin{itemize}
    \item We propose a novel method for unsupervised activity segmentation, which jointly performs representation learning and online clustering. We leverage video frame clustering as a pretext task, thus directly optimizing for unsupervised activity segmentation.
    \item We introduce the temporal optimal transport module to exploit temporal cues in videos by imposing temporal order-preserving constraints on computed pseudo-label cluster assignments, yielding effective representations for unsupervised activity segmentation.
    \item Our method performs on par with or better than the state-of-the-art in unsupervised activity segmentation on public datasets, i.e., 50-Salads, YouTube Instructions, and Breakfast, and our dataset, i.e., Desktop Assembly, while being much more memory efficient.
    \item We collect and label our Desktop Assembly dataset.
\end{itemize}
\section{Related Work}
\label{sec:relatedwork}

Below we summarize related works in temporal activity segmentation and self-supervised representation learning.

\noindent \textbf{Unsupervised Activity Segmentation.}
Early methods~\cite{malmaud2015s,sener2015unsupervised,alayrac2016unsupervised} in unsupervised activity segmentation explore cues from the accompanying narrations for segmenting the videos. They assume the narrations are available and well-aligned with the videos, which is not always the case and hence limits their applications. Approaches ~\cite{sener2018unsupervised,kukleva2019unsupervised,vidalmata2021joint,li2021action} which rely purely on visual inputs have been developed recently. Sener et al.~\cite{sener2018unsupervised} propose an iterative approach which alternates between learning a discriminative appearance model and optimizing a generative temporal model of the activity, while Kukleva et al.~\cite{kukleva2019unsupervised} introduce a multi-step approach which includes learning a temporal embedding and performing K-means clustering on the learned features. VidalMata et al.~\cite{vidalmata2021joint} and Li and Todorovic~\cite{li2021action} further improve the approach of~\cite{kukleva2019unsupervised} by learning a visual embedding and an action-level embedding respectively. The above approaches~\cite{sener2018unsupervised,kukleva2019unsupervised,vidalmata2021joint,li2021action} usually separate representation learning from clustering, and require storing learned features for the whole dataset before clustering them. In contrast, our approach combines representation learning and clustering into a single joint framework, while processing one mini-batch at a time, leading to better results and memory efficiency. More recently, the work by Swetha et al.~\cite{swetha2021unsupervised} proposes a joint representation learning and clustering approach. However, our approach is different from theirs in several aspects. Firstly, we employ optimal transport for clustering, while they use discriminative learning. Secondly, for representation learning, we employ clustering-based loss, while they use reconstruction loss. Lastly, despite our simpler encoder, our approach has similar or superior performance than theirs on public datasets.

\noindent \textbf{Weakly-Supervised Activity Segmentation.}
A few works focus on weak supervision for temporal activity segmentation such as the order of actions appearing in a video, i.e., transcript supervision~\cite{huang2016connectionist,richard2016temporal,kuehne2017weakly,richard2017weakly,richard2018neuralnetwork,ding2018weakly,chang2019d3tw,li2019weakly}, and the set of actions occurring in a video, i.e., set supervision~\cite{richard2018action,fayyaz2020sct,li2020set}. Recently, Li et al.~\cite{li2021temporal} apply timestamp supervision for temporal activity segmentation, 
which requires annotating a single frame for each action segment. Our approach, however, does not require any action labels.

\noindent \textbf{Image-Based Self-Supervised Representation Learning.}
Since the early work of Hinton and Zemel~\cite{hinton1994autoencoders}, considerable efforts~\cite{vincent2008extracting,larsson2016learning,larsson2017colorization,noroozi2017representation,liu2018leveraging,kim2018learning,gidaris2018unsupervised,carlucci2019domain,feng2019self} have been invested in designing pretext tasks with artificial image labels for training deep networks for self-supervised representation learning. These include image denoising~\cite{vincent2008extracting}, image colorization~\cite{larsson2016learning,larsson2017colorization}, object counting~\cite{noroozi2017representation,liu2018leveraging}, solving jigsaw puzzles~\cite{kim2018learning,carlucci2019domain}, and predicting image rotations~\cite{gidaris2018unsupervised,feng2019self}. Recently, a few approaches~\cite{bautista2016cliquecnn,xie2016unsupervised,yang2016joint,caron2018deep,caron2019unsupervised,asano2019self,huang2019unsupervised,zhuang2019local,caron2020unsupervised,gidaris2020learning,yan2020clusterfit} leveraging clustering as a pretext task have been introduced. For example, in~\cite{caron2018deep,caron2019unsupervised}, K-means cluster assignments are used as pseudo-labels for learning self-supervised image representations, while the pseudo-label assignments are obtained by solving the optimal transport problem in~\cite{asano2019self,caron2020unsupervised}. In this paper, we focus on learning self-supervised video representations, which requires exploring both spatial and temporal cues in videos. In particular, we follow the clustering-based approaches of~\cite{asano2019self,caron2020unsupervised}, however, unlike them, we employ temporal optimal transport to leverage temporal cues.

\noindent \textbf{Video-Based Self-Supervised Representation Learning.}
Over the past few decades, a variety of pretext tasks have been proposed for learning self-supervised video representations~\cite{hadsell2006dimensionality,mobahi2009deep,bengio2009slow,zou2011unsupervised,zou2012deep,srivastava2015unsupervised,goroshin2015unsupervised,vondrick2016generating,misra2016shuffle,lee2017unsupervised,fernando2017self,ahsan2018discrimnet,diba2019dynamonet,han2019video,kim2019self,gammulle2019predicting,xu2019self,choi2020shuffle}. A popular group of methods learn representations by predicting future frames~\cite{srivastava2015unsupervised,vondrick2016generating,ahsan2018discrimnet,diba2019dynamonet} or their encoding features~\cite{han2019video,kim2019self,gammulle2019predicting}. Another group explore temporal information such as temporal order~\cite{misra2016shuffle,lee2017unsupervised,fernando2017self, xu2019self,choi2020shuffle} and temporal coherence~\cite{hadsell2006dimensionality,mobahi2009deep,bengio2009slow,zou2011unsupervised,zou2012deep,goroshin2015unsupervised}. The above approaches process a single video at a time. Recently, a few methods~\cite{sermanet2018time,dwibedi2019temporal,haresh2021learning} which optimize over a pair of videos at once have been introduced. TCN~\cite{sermanet2018time} learns representations via the time-contrastive loss across different viewpoints and neighboring frames, while TCC~\cite{dwibedi2019temporal} and LAV~\cite{haresh2021learning} perform frame matching and temporal alignment between videos respectively. Here, we learn self-supervised representations by clustering video frames, which directly optimizes for the downstream task of unsupervised activity segmentation.
\section{Our Approach}
\label{sec:method}

\begin{figure*}[t]
	\centering
		\includegraphics[width=0.8\linewidth, trim = 0mm 85mm 95mm 0mm, clip]{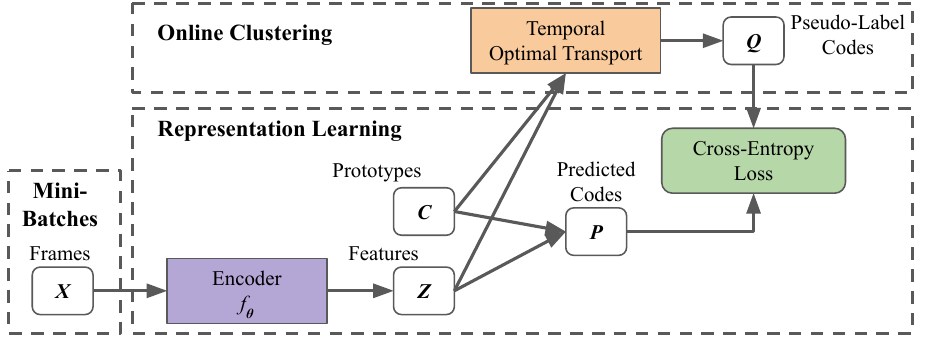}
	\caption{Given the frames $\boldsymbol{X}$, we feed them to the encoder $f_{\boldsymbol{\theta}}$ to obtain the features $\boldsymbol{Z}$, which are combined with the prototypes $\boldsymbol{C}$ to produce the predicted codes $\boldsymbol{P}$. Meanwhile, $\boldsymbol{Z}$ and $\boldsymbol{C}$ are also fed to the temporal optimal transport module to compute the pseudo-label codes $\boldsymbol{Q}$. We jointly learn $\boldsymbol{\theta}$ and $\boldsymbol{C}$ by applying the cross-entropy loss on $\boldsymbol{P}$ and $\boldsymbol{Q}$.}
	\label{fig:overview}
\end{figure*}

We now describe our main contribution, which is an unsupervised approach for activity segmentation. In particular, we propose a joint self-supervised representation learning and online clustering approach, which uses video frame clustering as a pretext task and hence directly optimizes for unsupervised activity segmentation. We exploit temporal information in videos by using temporal optimal transport. Fig.~\ref{fig:overview} shows an overview of our approach. Below we first define some notations and then provide the details of our representation learning and online clustering modules.

\noindent \textbf{Notations.}
We denote the embedding function as $f_{\boldsymbol{\theta}}$, i.e., a neural network with learnable parameters $\boldsymbol{\theta}$. Our approach takes as input a mini-batch $\boldsymbol{X} = \{\boldsymbol{x}_1, \boldsymbol{x}_2, \dots, \boldsymbol{x}_B\}$, where $B$ is the number of frames in $\boldsymbol{X}$. For a frame $\boldsymbol{x}_i$ in $\boldsymbol{X}$, the embedding features of $\boldsymbol{x}_i$ are expressed as $\boldsymbol{z}_i = f_{\boldsymbol{\theta}}(\boldsymbol{x}_i) \in \mathbb{R}^{D}$, with $D$ being the dimension of the embedding features. The embedding features of $\boldsymbol{X}$ are then written as $\boldsymbol{Z} = [\boldsymbol{z}_1, \boldsymbol{z}_2, \dots, \boldsymbol{z}_B]^\top \in \mathbb{R}^{B \times D}$. Moreover, we denote $\boldsymbol{C} = [\boldsymbol{c}_1, \boldsymbol{c}_2, \dots, \boldsymbol{c}_K]^\top \in \mathbb{R}^{K \times D}$ as the learnable prototypes of the $K$ clusters, with $\boldsymbol{c}_j$ representing the prototype of the $j$-th cluster. Lastly, $\boldsymbol{P} \in \mathbb{R}_{+}^{B \times K}$ and $\boldsymbol{Q} \in \mathbb{R}_{+}^{B \times K}$ are the predicted cluster assignments (i.e., predicted ``codes'') and pseudo-label cluster assignments (i.e., pseudo-label ``codes'') respectively.

\subsection{Representation Learning}
\label{sec:representation}

To learn self-supervised representations for unsupervised activity segmentation, our proposed idea is to use video frame clustering as a pretext task. Thus, the learned features are explicitly optimized for unsupervised activity segmentation. Here, we consider a similar clustering-based self-supervised representation learning approach as~\cite{asano2019self,caron2020unsupervised}. However, unlike their approaches which are designed for image data, we propose temporal optimal transport to make use of temporal information additionally available in video data. Below we describe our losses for learning representations for unsupervised activity segmentation.

\noindent \textbf{Cross-Entropy Loss.}
Given the frames $\boldsymbol{X}$, we first pass them to the encoder $f_{\boldsymbol{\theta}}$ to obtain the features $\boldsymbol{Z}$. We then compute the predicted codes $\boldsymbol{P}$ with each entry written as:
\begin{align}
\boldsymbol{P}_{ij} = \frac{ \exp ( \frac{1}{\tau} \boldsymbol{z}^\top_i \boldsymbol{c}_j ) } { \sum^{K}_{j'=1} \exp ( \frac{1}{\tau} \boldsymbol{z}^\top_i \boldsymbol{c}_{j'} ) },
\end{align}
where $\boldsymbol{P}_{ij}$ is the probability that the $i$-th frame is assigned to the $j$-th cluster and $\tau$ is the temperature parameter~\cite{wu2018unsupervised}. The pseudo-label codes $\boldsymbol{Q}$ are computed by solving the temporal optimal transport problem, which we will describe in the next section. For clustering-based representation learning, we minimize the cross-entropy loss with respect to the encoder parameters $\boldsymbol{\theta}$ and the prototypes $\boldsymbol{C}$ as:
\begin{align}
\label{eq:cel}
    L_{CE} = - \frac{1}{B} \sum^{B}_{i=1} \sum^{K}_{j=1} \boldsymbol{Q}_{ij} \log \boldsymbol{P}_{ij}.
\end{align}

\noindent \textbf{Temporal Coherence Loss.}
To further exploit temporal information in videos, we consider adding another self-supervised loss, i.e., the temporal coherence loss. It learns an embedding space following the temporal coherence constraints~\cite{hadsell2006dimensionality,mobahi2009deep,goroshin2015unsupervised}, where temporally close frames should be mapped to nearby points and temporally distant frames should be mapped to far away points. To enable fast convergence and effective representations, we employ the N-pair metric learning loss proposed by~\cite{sohn2016improved}. For each video, we first sample a subset of $N$ ordered frames denoted by $\{\boldsymbol{z}_i\}$ (with $i \in \{1, 2, ..., N\}$). For each $\boldsymbol{z}_i$, we then sample a ``positive'' example $\boldsymbol{z}^+_i$ inside a temporal window of $\lambda$ from $\boldsymbol{z}_i$. Moreover, $\boldsymbol{z}^+_j$ sampled for $\boldsymbol{z}_j$ (with $j\neq i$) is considered as a ``negative'' example for $\boldsymbol{z}_i$. We minimize the temporal coherence loss with respect to the encoder parameters $\boldsymbol{\theta}$ as:
\begin{align}
\label{eq:tcl}
    L_{TC} = - \frac{1}{N} \sum^{N}_{i=1} \log \frac{\exp(\boldsymbol{z}^\top_i \boldsymbol{z}^+_i)}{ \sum_{j=1}^{N} \exp(\boldsymbol{z}^{\top}_i \boldsymbol{z}^+_j)}.
\end{align}

\noindent \textbf{Final Loss.}
Our final loss is written as:
\begin{align}
\label{eq:finalloss}
    L = L_{CE} + \alpha L_{TC}.
\end{align}
Here, $\alpha$ is the weight for the temporal coherence loss. Our final loss is optimized with respect to $\boldsymbol{\theta}$ and $\boldsymbol{C}$. The cross-entropy loss and the temporal coherence loss are differentiable and can be optimized using backpropagation. Note that we do not backpropagate through $\boldsymbol{Q}$.

\subsection{Online Clustering}
\label{sec:online}

Below we describe our online clustering module for computing the pseudo-label codes $\boldsymbol{Q}$ online. Following~\cite{asano2019self,caron2020unsupervised}, we consider the problem of computing $\boldsymbol{Q}$ as the optimal transport problem and solve for $\boldsymbol{Q}$ online by using a mini-batch $\boldsymbol{X}$ at a time. This is different from prior works~\cite{sener2018unsupervised,kukleva2019unsupervised,vidalmata2021joint,li2021action} for unsupervised activity segmentation, which require storing features for the entire dataset before clustering them in an offline fashion and hence have significantly more memory constraints.

\noindent \textbf{Optimal Transport.}
Given the features $\boldsymbol{Z}$ extracted from the frames $\boldsymbol{X}$, our goal is to compute the pseudo-label codes $\boldsymbol{Q}$ with each entry $\boldsymbol{Q}_{ij}$ representing the probability that the features $\boldsymbol{z}_i$ are mapped to the prototype $\boldsymbol{c}_j$. Specifically, $\boldsymbol{Q}$ is computed by solving the optimal transport problem as:
\begin{align}
    \label{eq:ot}
    \max_{\boldsymbol{Q} \in \mathcal{Q}}~~~ Tr(\boldsymbol{Q}^\top \boldsymbol{Z} \boldsymbol{C}^\top) + \epsilon H(\boldsymbol{Q}),
\end{align}
\begin{align}
    \label{eq:equipartition}
    \mathcal{Q} = \left\{ \boldsymbol{Q} \in \mathbb{R}_{+}^{B \times K}:~~~ \boldsymbol{Q}\boldsymbol{1}_K = \frac{1}{B} \boldsymbol{1}_B, \boldsymbol{Q}^\top \boldsymbol{1}_B = \frac{1}{K} \boldsymbol{1}_K \right\}.
\end{align}
Here, $\boldsymbol{1}_B$ and $\boldsymbol{1}_K$ denote vectors of ones in dimensions $B$ and $K$ respectively. In Eq.~\ref{eq:ot}, the first term measures the similarity between the features $\boldsymbol{Z}$ and the prototypes $\boldsymbol{C}$, while the second term (i.e., $H(\boldsymbol{Q}) = - \sum^{B}_{i=1} \sum^{K}_{j=1} \boldsymbol{Q}_{ij} \log \boldsymbol{Q}_{ij}$) measures the entropy regularization of $\boldsymbol{Q}$, and $\epsilon$ is the weight for the entropy term. A large value of $\epsilon$ usually leads to a trivial solution where every frame has the same probability of being assigned to every cluster. Thus, we use a small value of $\epsilon$ in our experiments to avoid the above trivial solution. Furthermore, Eq.~\ref{eq:equipartition} represents the \emph{equal partition} constraints, which enforce that each cluster is assigned the same number of frames in a mini-batch, thus preventing a trivial solution where all frames are assigned to a single cluster. Although the above equal partition prior does not hold for activities with various action lengths, we find that in practice it works relatively well for most activities with various action lengths (e.g., please see Fig.~\ref{fig:segs} and more discussion in the supplementary material). The solution for the above optimal transport problem can be computed by using the iterative Sinkhorn-Knopp algorithm~\cite{cuturi2013sinkhorn} as:
\begin{align}
    \boldsymbol{Q}_{OT} = \diag(\boldsymbol{u}) \exp \left( \frac{\boldsymbol{Z} \boldsymbol{C}^\top}{\epsilon} \right) \diag(\boldsymbol{v}),
\end{align}
where $\boldsymbol{u} \in \mathbb{R}^{B}$ and $\boldsymbol{v} \in \mathbb{R}^{K}$ are renormalization vectors.

\noindent \textbf{Temporal Optimal Transport.}
The above approach is originally developed for image data in~\cite{asano2019self,caron2020unsupervised} and hence is not capable of exploiting temporal cues in video data for unsupervised activity segmentation. Thus, we propose to incorporate a temporal regularization term which preserves the temporal order of the activity into the objective in Eq.~\ref{eq:ot}, yielding the temporal optimal transport.

Motivated by~\cite{su2017order}, we introduce a prior distribution for $\boldsymbol{Q}$, namely $\boldsymbol{T} \in \mathbb{R}_{+}^{B \times K}$, where the highest values appear on the diagonal and the values gradually decrease along the direction perpendicular to the diagonal. Specifically, $\boldsymbol{T}$ maintains a \emph{fixed order} of the clusters, and enforces initial frames to be assigned to initial clusters and later frames to be assigned to later clusters. Mathematically, $\boldsymbol{T}$ can be represented by a 2D distribution, whose marginal distribution along any line perpendicular to the diagonal is a Gaussian distribution centered at the intersection on the diagonal, as:
\begin{align}
    \boldsymbol{T}_{ij} = \frac{1}{\sigma \sqrt{2 \pi}} \exp\left(- \frac{d_{ij}^2}{2 \sigma^2}\right), d_{ij} = \frac{|i/B - j/K|}{\sqrt{1/B^2 + 1/K^2}},
\end{align}
where $d_{ij}$ is the distance from the entry $(i,j)$ to the diagonal line. Though the above temporal order-preserving prior does not hold for activities with permutations, we empirically observe that it performs relatively well on most datasets containing permutations (e.g., please see Tabs.~\ref{tab:50Salds_results}, \ref{tab:yti_results}, \ref{tab:breakfast_results}, and more discussion in the supplementary material).

To encourage the distribution of values of $\boldsymbol{Q}$ to be as similar as possible to $\boldsymbol{T}$, we replace the objective in Eq.~\ref{eq:ot} with the temporal optimal transport objective:
\begin{align}
    \label{eq:tot}
    \max_{\boldsymbol{Q} \in \mathcal{Q}}~~~ Tr(\boldsymbol{Q}^\top \boldsymbol{Z} \boldsymbol{C}^\top) - \rho KL(\boldsymbol{Q}||\boldsymbol{T}).
\end{align}
Here, $KL(\boldsymbol{Q}||\boldsymbol{T}) = \sum^{B}_{i=1} \sum^{K}_{j=1} \boldsymbol{Q}_{ij} \log \frac{\boldsymbol{Q}_{ij}}{\boldsymbol{T}_{ij}}$ is the Kullback-Leibler (KL) divergence between $\boldsymbol{Q}$ and $\boldsymbol{T}$, and $\rho$ is the weight for the KL term. Note that $\mathcal{Q}$ is defined as in Eq.~\ref{eq:equipartition}. Following~\cite{cuturi2013sinkhorn}, we can derive the solution for the above temporal optimal transport problem as:
\begin{align}
    \boldsymbol{Q}_{TOT} = \diag(\boldsymbol{u}) \exp \left( \frac{\boldsymbol{Z} \boldsymbol{C}^\top + \rho \log \boldsymbol{T}}{\rho} \right) \diag(\boldsymbol{v}),
\end{align}
where $\boldsymbol{u} \in \mathbb{R}^{B}$ and $\boldsymbol{v} \in \mathbb{R}^{K}$ are renormalization vectors. 

In contrast to previous methods \cite{li2021action, vidalmata2021joint, kukleva2019unsupervised, sener2018unsupervised} which require features of the entire dataset to be loaded into memory, our method requires only a mini-batch of features to be loaded in memory at a time. This reduces the memory requirement significantly from $O(N)$ to $O(B)$, where $B$ is the mini-batch size, $N$ is the total number of frames in the entire dataset, and $B$ is much smaller than $N$, especially for large datasets. For example, CTE \cite{kukleva2019unsupervised} requires a memory of $57795 \times 30 \times 8$ bytes for storing features on the 50 Salads dataset, whereas our method requires $512 \times 30 \times 8$ bytes for the same purpose, where $N = 57795$, $B = 512$, and 30 is the size of the final embedding.
\section{Experiments}
\label{sec:experiments}

\noindent \textbf{Implementation Details.} We use a 2-layer MLP for learning the embedding on top of pre-computed features (see below). The MLP is followed by a dot-product operation with the prototypes which are initialized randomly and learned via backpropagation through the losses presented in Sec.~\ref{sec:representation}. The ADAM optimizer~\cite{kingma2014adam} is used with a learning rate of $10^{-3}$ and a weight decay of $10^{-4}$. For each activity, the number of prototypes is set as the number of actions in the activity. For our approach, the order of the actions is fixed as mentioned in Sec.~\ref{sec:online}. During inference, cluster assignment probabilities for all frames are computed. These probabilities are then passed to a Viterbi decoder for smoothing out the probabilities given the order of the actions. Note that, for a fair comparison, the above protocol is the same as in CTE~\cite{kukleva2019unsupervised}, which is the closest work to ours. Please see more details in the supplementary material.

\noindent \textbf{Datasets.} We use three public datasets (all under Creative Commons License), namely 50 Salads~\cite{stein2013combining}, YouTube Instructions (YTI)~\cite{alayrac2016unsupervised}, and Breakfast~\cite{kuehne2014language}, while introducing our Desktop Assembly dataset:
\begin{itemize}
    \item \emph{50 Salads} consists of $50$ videos of actors performing a cooking activity. The total video duration is about $4.5$ hours. Following previous works, we report results at two granularity levels, i.e., \emph{Eval} with $12$ action classes and \emph{Mid} with $19$ action classes. For \emph{Eval}, some action classes are merged into one class (e.g., ``cut cucumber'', ``cut tomato'', and ``cut cheese'' are  all considered as ``cut''). Thus, it has less number of action classes than \emph{Mid}. We use pre-computed features by~\cite{wang2013action}.
    \item \emph{YouTube Instructions (YTI)} includes 150 videos belonging to 5 activities. The average video length is about 2 minutes. This dataset also has a large number of frames labeled as background. Following previous works, we use pre-computed features provided by~\cite{alayrac2016unsupervised}.
    \item \emph{Breakfast} consists of 10 activities with about 8 actions per activity. The average video length varies from few seconds to several minutes depending on the activity. Following previous works, we use pre-computed features proposed by~\cite{kuehne2016end} and shared by~\cite{kukleva2019unsupervised}.
    \item Our \emph{Desktop Assembly} dataset includes 76 videos of actors performing an assembly activity. The activity comprises 22 actions conducted in a fixed order. Each video is about 1.5 minutes long. We use pre-computed features from ResNet-18~\cite{he2016deep} pre-trained on ImageNet. Please see more details in the supplementary material.
\end{itemize}

\noindent \textbf{Metrics.} Since no labels are provided for training, there is no direct mapping between predicted and ground truth segments. To establish this mapping, we follow~\cite{sener2018unsupervised, kukleva2019unsupervised} and perform Hungarian matching. Note that the Hungarian matching is conducted at the activity level, i.e., it is computed over all frames of an activity. This is different from the Hungarian matching used in~\cite{aakur2019perceptual} which is done at the video level and generally leads to better results due to more fine-grained matching~\cite{vidalmata2021joint}. We adopt Mean Over Frames (MOF) and F1-Score as our metrics. MOF is the percentage of correct frame-wise predictions averaged over all activities. For F1-Score, to compute precision and recall, positive detections must have more than 50\% overlap with ground truth segments. F1-Score is computed for each video and averaged over all videos. Please see~\cite{kukleva2019unsupervised} for more details.

\noindent \textbf{Competing Methods.} We compare against various unsupervised activity segmentation methods~\cite{alayrac2016unsupervised,sener2018unsupervised,kukleva2019unsupervised,vidalmata2021joint,li2021action,swetha2021unsupervised}. Frank-Wolfe~\cite{alayrac2016unsupervised} explores accompanied narrations. Mallow~\cite{sener2018unsupervised} iterates between representation learning based on discriminative learning and temporal modeling based on a generalized Mallows model. CTE~\cite{kukleva2019unsupervised} leverages time-stamp prediction for representation learning and then K-means for clustering. VTE~\cite{vidalmata2021joint} and ASAL~\cite{li2021action} further improve CTE~\cite{kukleva2019unsupervised} with visual cues (via future frame prediction) and action-level cues (via action shuffle prediction) respectively. UDE~\cite{swetha2021unsupervised} uses discriminative learning for clustering and reconstruction loss for representation learning.

\subsection{Ablation Study Results}
\label{sec:ablation}

\begin{figure}[t]
     \centering
     \begin{subfigure}[b]{0.22\textwidth}
         \centering
         \includegraphics[width=\textwidth]{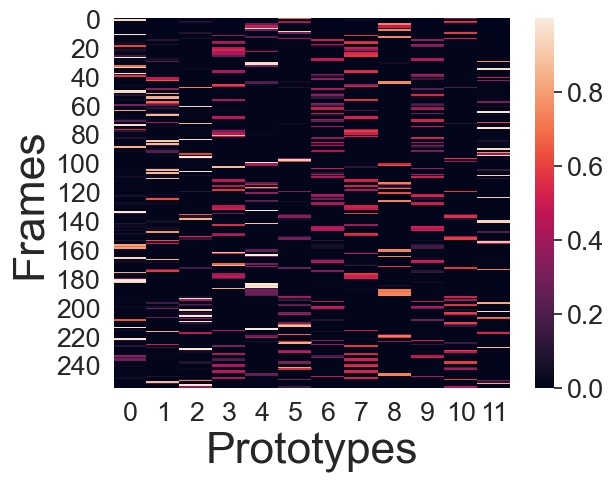}
         \caption{OT}
         \label{fig:q_mat_ot}
     \end{subfigure}
     \hfill
     \centering
     \begin{subfigure}[b]{0.22\textwidth}
         \centering
         \includegraphics[width=\textwidth]{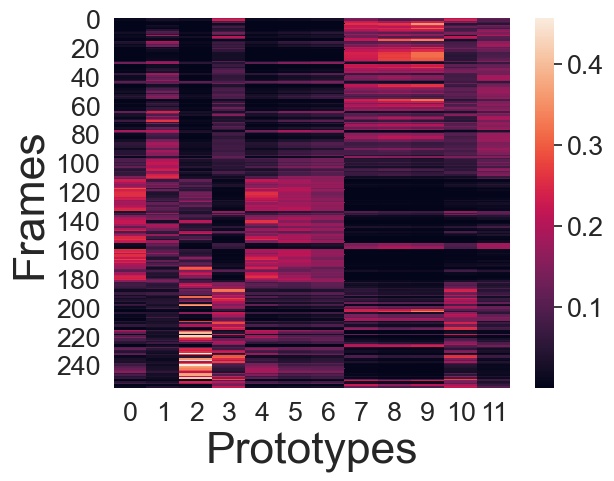}
         \caption{OT+CTE}
         \label{fig:q_mat_ot}
     \end{subfigure}
     \hfill
     \begin{subfigure}[b]{0.22\textwidth}
         \centering
         \includegraphics[width=\textwidth]{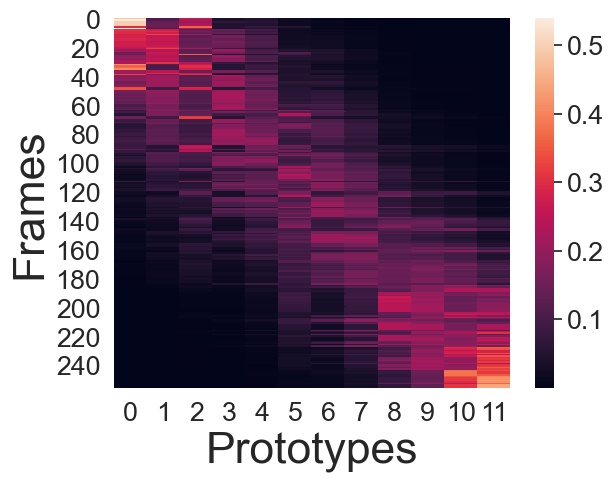}
         \caption{TOT}
         \label{fig:q_mat_tot}
     \end{subfigure}
     \hfill
     \begin{subfigure}[b]{0.22\textwidth}
         \centering
         \includegraphics[width=\textwidth]{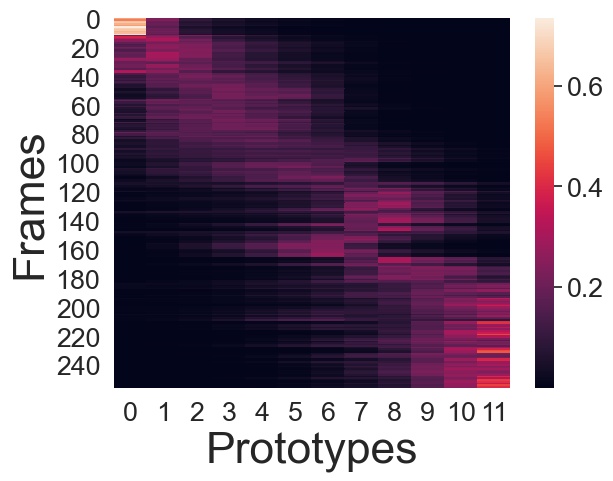}
         \caption{TOT+TCL}
         \label{fig:q_mat_tot}
     \end{subfigure}
    \caption{Pseudo-label codes $\boldsymbol{Q}$ computed by different variants for a 50 Salads video.}
    \label{fig:q_mat}
\end{figure}

We perform ablation studies on 50 Salads (i.e., \emph{Eval} granularity) and YTI to show the effectiveness of our design choices in Sec.~\ref{sec:method}. Tabs.~\ref{tab:ablation_50salads} and~\ref{tab:ablation_yti} show the ablation study results. We first begin with the standard optimal transport (OT), without any temporal prior. From Tabs.~\ref{tab:ablation_50salads} and~\ref{tab:ablation_yti}, OT has the worst overall performance, e.g., OT obtains $27.8$ for F1-Score on 50 Salads, and $11.6$ for F1-Score and $16.0\%$ for MOF on YTI. Next, we experiment with adding temporal priors to OT, including time-stamp prediction loss in CTE~\cite{kukleva2019unsupervised} (yielding \emph{OT+CTE}), temporal coherence loss in Sec.~\ref{sec:representation} (yielding \emph{OT+TCL}), and temporal order-preserving prior in Sec.~\ref{sec:online} (yielding \emph{TOT}). We notice while OT+CTE, OT+TCL, and TOT all outperform OT, TOT achieves the best performance among them, e.g., TOT obtains $42.8$ for F1-Score on 50 Salads, and $30.0$ for F1-Score and $40.6\%$ for MOF on YTI. The above observations are also confirmed by plotting the pseudo-label codes $\boldsymbol{Q}$ computed by different variants in Fig.~\ref{fig:q_mat}. It can be seen that OT fails to capture any temporal structure of the activity, whereas TOT manages to capture the temporal order of the activity relatively well (i.e., initial frames should be mapped to initial prototypes and vice versa).

Finally, we consider adding more temporal priors to TOT, including time-stamp prediction loss in CTE~\cite{kukleva2019unsupervised} (yielding \emph{TOT+CTE}) and temporal coherence loss in Sec.~\ref{sec:representation} (yielding \emph{TOT+TCL}). We observe that TCL is often complementary to TOT, and TOT+TCL achieves the best overall performance, e.g., TOT+TCL obtains $48.2$ for F1-Score on 50 Salads, and $32.9$ for F1-Score and $45.3\%$ for MOF on YTI. We notice that TOT+TCL has a lower MOF than TOT on 50 Salads, which might be because TCL optimizes for disparate representations for different actions but multiple action classes are merged into one in 50 Salads (i.e., \emph{Eval} granularity).
\begin{table}[t]
\begin{minipage}[c]{1.0\linewidth}
\centering
\begin{tabular}{l|l|l}

\specialrule{1pt}{1pt}{1pt}

\textbf{Variants}  &  \textbf{F1-Score} & \textbf{MOF}\\

\midrule
 OT	& 27.8 &	37.6 \\
 OT+CTE & 34.3 &	40.4 \\
 OT+TCL &	30.3 &	27.5 \\
\rowcolor{beaublue}  TOT &	\textit{\underline{42.8}} &	\textbf{47.4} \\
 TOT+CTE &	36.0 &	40.8 \\
\rowcolor{beaublue}   TOT+TCL &	\textbf{48.2} &	\textit{\underline{44.5}} 

\\

\specialrule{1pt}{1pt}{1pt}
\end{tabular}
\caption{Ablation study results on 50 Salads (i.e., \emph{Eval} granularity). The best results are in \textbf{bold}. The second best are \textit{\underline{underlined}}.}
\label{tab:ablation_50salads}
\end{minipage}

\begin{minipage}[c]{1.0\linewidth}
\centering
\begin{tabular}{l|l|l}

\specialrule{1pt}{1pt}{1pt}

\textbf{Variants}  &  \textbf{F1-Score} & \textbf{MOF}\\

\midrule

OT & 11.6 &	16.0 \\
OT+CTE &	22.0 &	35.2 \\
OT+TCL &	24.8 & 35.7 \\
\rowcolor{beaublue} TOT &	\textit{\underline{30.0}} &	\textit{\underline{40.6}} \\
TOT+CTE &	26.7 &	38.2 \\
\rowcolor{beaublue} TOT+TCL &	\textbf{32.9} &	\textbf{45.3} \\

\specialrule{1pt}{1pt}{1pt}
\end{tabular}
\caption{Ablation study results on YouTube Instructions. The best results are in \textbf{bold}. The second best are \textit{\underline{underlined}}.}
\label{tab:ablation_yti}
\end{minipage}
\end{table}

\subsection{Hyperparameter Setting Results}
\label{sec:ablation}

\begin{figure}[t]
     \centering
     \begin{subfigure}[b]{0.22\textwidth}
         \centering
         \includegraphics[width=\textwidth]{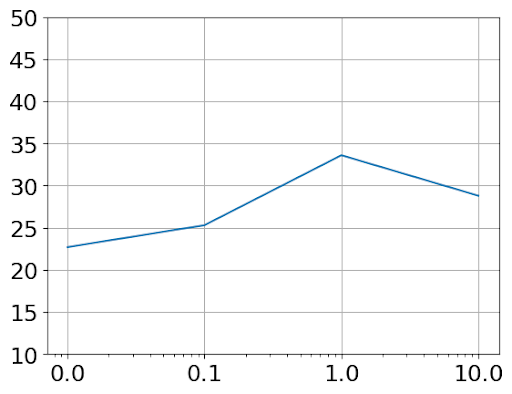}
         \caption{$\alpha$}
         \label{fig:sen_alpha}
     \end{subfigure}
     \hfill
     \centering
     \begin{subfigure}[b]{0.22\textwidth}
         \centering
         \includegraphics[width=\textwidth]{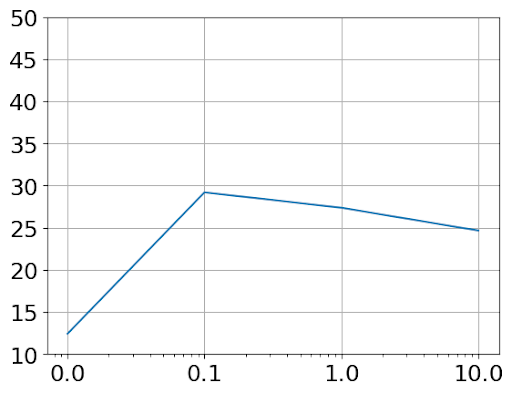}
         \caption{$\rho$}
         \label{fig:sen_rho}
     \end{subfigure}
     \hfill
     \begin{subfigure}[b]{0.22\textwidth}
         \centering
         \includegraphics[width=\textwidth]{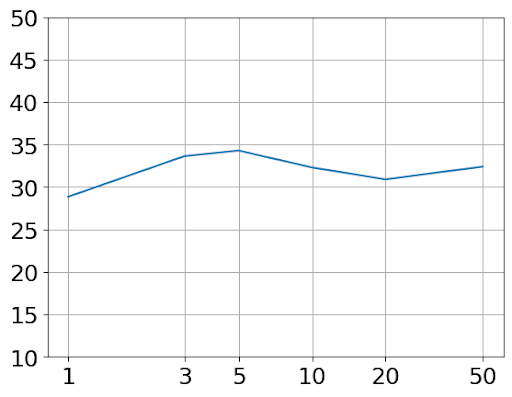}
         \caption{$\eta$}
         \label{fig:sen_sink}
     \end{subfigure}
     \hfill
     \begin{subfigure}[b]{0.22\textwidth}
         \centering
         \includegraphics[width=\textwidth]{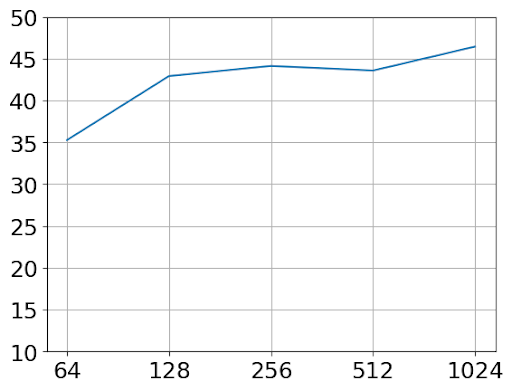}
         \caption{$B$}
         \label{fig:sen_bs}
     \end{subfigure}
    \caption{Hyperparameter setting results. Y axes show F1-Scores. We use YTI in (a-c) and 50 Salads (\emph{Eval} granularity) in (d).}
    \label{fig:ablation}
\end{figure}

\noindent\textbf{Effects of $\alpha$.} We study the effects of different values of $\alpha$, i.e., the balancing weight between the clustering-based loss and the temporal coherence loss in Eq.~\ref{eq:finalloss}. We measure F1-Scores on YouTube Instructions. Fig.~\ref{fig:ablation}(a) shows the results, where the performance peaks in the proximity of $\alpha = 1.0$.

\noindent\textbf{Effects of $\rho$.} The effects of various values of $\rho$, i.e., the balancing weight between the similarity term and the temporal order-preserving term in Eq.~\ref{eq:tot}, are presented in Fig.~\ref{fig:ablation}(b). We use YouTube Instructions and measure F1-Scores. From Fig.~\ref{fig:ablation}(b), $\rho \in  [0.07,0.1]$ performs the best. The drop at $\rho = 0.01$ is due to numerical issues (see Fig.~6 of~\cite{su2017order}).

\noindent\textbf{Effects of $\eta$.} Fig.~\ref{fig:ablation}(c) shows the results of varying the value of $\eta$, i.e., the number of Sinkhorn-Knopp iterations during TOT training. We measure F1-Scores on YouTube Instructions. From the results, $\eta \in [3, 5]$ performs the best. Larger values of $\eta$ do not improve the performance but increase the computational cost significantly.

\noindent\textbf{Effects of $B$.} The results of increasing the value of $B$, i.e., the mini-batch size during TOT training, are presented in Fig.~\ref{fig:ablation}(d). We use 50 Salads dataset (\emph{Eval} granularity) and measure F1-Scores. As we can see from the results, the performance improves as the mini-batch size increases.

\subsection{Results on 50 Salads Dataset}
\label{sec:50salads}

Tab.~\ref{tab:50Salds_results} presents the MOF results of different unsupervised activity segmentation methods on 50 Salads. From the results, TOT outperforms CTE~\cite{kukleva2019unsupervised} by $11.9\%$ and $1.6\%$ on the \emph{Eval} and \emph{Mid} granularity respectively. Similarly, TOT also outperforms VTE~\cite{vidalmata2021joint} by $16.8\%$ and $7.6\%$ on the \emph{Eval} and \emph{Mid} granularity respectively. Note that CTE, which uses a sequential representation learning and clustering framework, is our most relevant competitor. VTE further improves CTE by exploring visual information via future frame prediction, which is not utilized in TOT. The significant performance gains of TOT over both CTE and VTE show the advantages of joint representation learning and clustering. Moreover, TOT performs the best on the \emph{Eval} granularity, outperforming the recent works of ASAL~\cite{li2021action} and UDE~\cite{swetha2021unsupervised} by $8.2\%$ and $5.2\%$  respectively. Finally, by combining TOT and TCL, we achieve $34.3\%$ on the \emph{Mid} granularity, which is very close to the best performance of $34.4\%$ of ASAL. Also, TOT+TCL outperforms ASAL and UDE by $5.3\%$ and $2.3\%$ on the \emph{Eval} granularity respectively. As mentioned previously, TOT+TCL has a lower MOF than TOT on the \emph{Eval} granularity, which might be due to large intra-class variations in the \emph{Eval} granularity.

\subsection{Results on YouTube Instructions Dataset}
\label{sec:yti}

Here, we compare our approach against state-of-the-art methods~\cite{alayrac2016unsupervised,sener2018unsupervised,kukleva2019unsupervised,vidalmata2021joint,li2021action,swetha2021unsupervised} for unsupervised activity segmentation on YTI. Following all of the above works, we report the performance without considering background frames. Tab.~\ref{tab:yti_results} presents the results. As we can see from Tab.~\ref{tab:yti_results}, TOT+TCL achieves the best performance on both metrics, outperforming all competing methods including the recent works of ASAL~\cite{li2021action} and UDE~\cite{swetha2021unsupervised}. In particular, TOT+TCL achieves $32.9$ for F1-Score, while ASAL and UDE obtain $32.1$ and $29.6$ respectively. Similarly, TOT+TCL achieves $45.3\%$ for MOF, while ASAL and UDE obtain $44.9\%$ and $43.8\%$ respectively. Finally, although TOT is inferior to TOT+TCL on both metrics, TOT outperforms a few competing methods. Specifically, TOT has a higher F1-Score than  UDE~\cite{swetha2021unsupervised}, VTE~\cite{vidalmata2021joint}, CTE~\cite{kukleva2019unsupervised}, Mallow~\cite{sener2018unsupervised}, and Frank-Wolfe~\cite{alayrac2016unsupervised}, and a higher MOF than CTE~\cite{kukleva2019unsupervised} and Mallow~\cite{sener2018unsupervised}.

\begin{table}[t]
\begin{minipage}{1.0\linewidth}
\centering {
\begin{tabular}{l|l|l}
\specialrule{1pt}{1pt}{1pt}
\textbf{Approach}  & \textbf{Eval} & \textbf{Mid}\\
\midrule
  CTE~\cite{kukleva2019unsupervised}
 
&  35.5 & 30.2   \\

  VTE~\cite{vidalmata2021joint}
 
&   30.6 & 24.2  \\

  ASAL~\cite{li2021action}
 
&    39.2 & \textbf{34.4} \\

 UDE~\cite{swetha2021unsupervised}
& 42.2 & - \\

\rowcolor{beaublue} Ours (TOT)
 
&  \textbf{47.4} & {31.8}  \\

\rowcolor{beaublue} Ours (TOT+TCL)

& \textit{\underline{44.5}}   & \textit{\underline{34.3}} \\

\specialrule{1pt}{1pt}{1pt}
\end{tabular}
}
\caption{Results on 50 Salads. The best results are in \textbf{bold}. The second best are \textit{\underline{underlined}}.}
\label{tab:50Salds_results}
\end{minipage}

\begin{minipage}{1.0\linewidth}
\centering
{
\begin{tabular}{l|l|l}
\specialrule{1pt}{1pt}{1pt}
  \textbf{Approach} &  \textbf{F1-Score} & \textbf{MOF}\\
\midrule
  Frank-Wolfe~\cite{alayrac2016unsupervised}
&   24.4 & -   \\

  Mallow~\cite{sener2018unsupervised}
 
&   27.0 & 27.8  \\

 CTE~\cite{kukleva2019unsupervised}

& 28.3 & 39.0 \\

  VTE~\cite{vidalmata2021joint}
 
&   29.9 & -  \\

  ASAL~\cite{li2021action}
 
&  \textit{\underline{32.1}} & \textit{\underline{44.9}} \\

 UDE~\cite{swetha2021unsupervised}

& 29.6 & 43.8 \\

\rowcolor{beaublue} Ours (TOT)
 
&  {30.0}    & {40.6}  \\

\rowcolor{beaublue} Ours (TOT+TCL)
 
& \textbf{32.9}    & \textbf{45.3}  \\

\specialrule{1pt}{1pt}{1pt}
\end{tabular}
}
\caption{Results on YouTube Instructions. The best results are in \textbf{bold}. The second best are \textit{\underline{underlined}}.}
\label{tab:yti_results}
\end{minipage}

\end{table}

\subsection{Results on Breakfast Dataset}
\label{sec:breakfast}
\begin{figure}
\begin{minipage}[c]{1.0\linewidth}
\captionsetup{type=table}
\centering
{
\begin{tabular}{l|l|l}

\specialrule{1pt}{1pt}{1pt}

\textbf{Approach} &  \textbf{F1-Score} & \textbf{MOF}\\

\midrule
Mallow~\cite{sener2018unsupervised} 
 
& - & 34.6 \\

CTE~\cite{kukleva2019unsupervised}

& 26.4 & 41.8 \\

VTE~\cite{vidalmata2021joint}
 
& - & \textit{\underline{48.1}} \\

ASAL~\cite{li2021action}
 
& \textbf{37.9}  & \textbf{52.5}  \\

UDE~\cite{swetha2021unsupervised}
 
& \textit{\underline{31.9}}  & 47.4  \\

\rowcolor{beaublue} Ours (TOT)
& 31.0 & 47.5 \\

\rowcolor{beaublue} Ours (TOT+TCL)

& 30.3 & 39.0 \\

\specialrule{1pt}{1pt}{1pt}
\end{tabular}
}
\caption{Results on Breakfast. The best results are in \textbf{bold}. The second best are \textit{\underline{underlined}}.}
\label{tab:breakfast_results}
\end{minipage}

\begin{minipage}[c]{1.0\linewidth}
    \captionsetup{type=figure} 
     \centering
     \includegraphics[width=\textwidth, trim = 0mm 5mm 10mm 0mm]{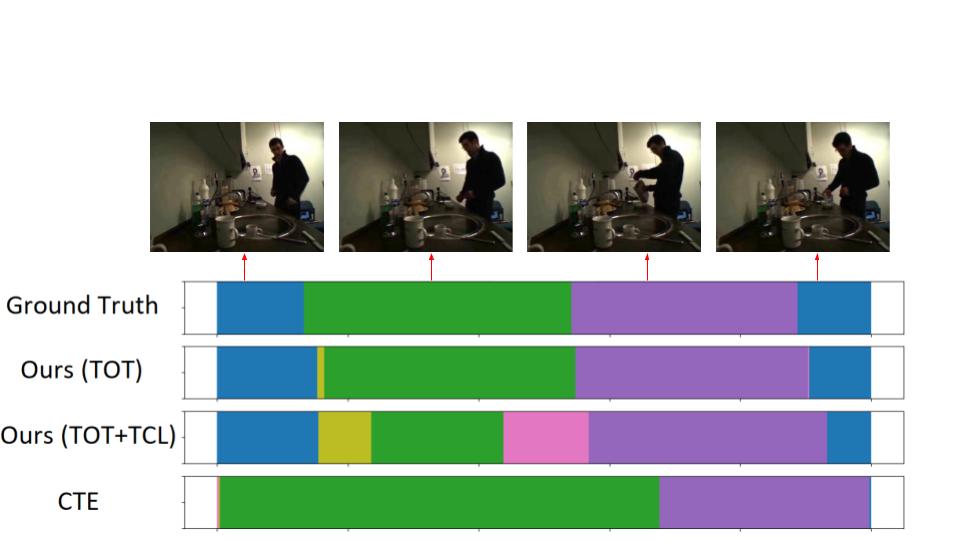}
    \caption{Segmentation results for a Breakfast video.}
    \label{fig:segs}
\end{minipage}
\end{figure}

We now discuss the performance of different methods on Breakfast. Tab.~\ref{tab:breakfast_results} shows the results. It can be seen that the recent work of ASAL~\cite{li2021action} obtains the best performance on both metrics. ASAL~\cite{li2021action} employs CTE~\cite{kukleva2019unsupervised} for initialization, and explores action-level cues for improvement, which can also be incorporated for boosting the performance of our approach. Next, TOT outperforms the sequential representation learning and clustering approach of CTE~\cite{kukleva2019unsupervised} by $4.6$ and $5.7\%$ on F1-Score and MOF respectively, while performing on par with VTE~\cite{vidalmata2021joint} and UDE~\cite{swetha2021unsupervised}, e.g., for MOF, TOT achieves $47.5\%$ while VTE and UDE obtain $48.1\%$ and $47.4\%$ respectively. Also, the significant performance gains of TOT over the most relevant competitor CTE confirms the advantages of joint representation learning and clustering. Some qualitative results are shown in Fig.~\ref{fig:segs}. It can be seen that our results are more closely aligned with the ground truth than those of CTE. Finally, combining TOT and TCL yields a similar F1-Score but a lower MOF than TOT, which might be due to large intra-class variations in the Breakfast dataset.

\subsection{Results on Desktop Assembly Dataset}
\label{sec:desktop}

Prior works, e.g., CTE~\cite{kukleva2019unsupervised} and VTE~\cite{vidalmata2021joint}, often exploit temporal information via time-stamp prediction. However, the same action might occur at various time stamps across videos in practice, e.g., different actors might perform the same action at different speeds. Our approach instead leverages temporal cues via temporal optimal transport, which preserves the temporal order of the activity. Tab.~\ref{tab:da_results} shows the results of CTE and our methods (i.e., TOT and TOT+TCL) on Desktop Assembly, where the activity comprises 22 actions conducted in a fixed order. From Tab.~\ref{tab:da_results}, TOT+TCL performs the best on both metrics, i.e., $53.4$ for F1-Score and $58.1\%$ for MOF. Also, TOT and TOT+TCL significantly outperform CTE on both metrics, i.e., TOT and TOT+TCL obtain F1-Score gains of $6.8$ and $8.5$ over CTE respectively, and MOF gains of $8.7\%$ and $10.5\%$ over CTE respectively.

\subsection{Generalization Results}
\label{sec:generalization}

So far, we have followed all previous works in unsupervised activity segmentation to use the same set of unlabelled videos for training and testing. We now explore another experiment setup to evaluate the generalization capability of our method. Specifically, we split the datasets, i.e., 50 Salads (\emph{Eval} granularity), YouTube Instructions, Breakfast, and Desktop Assembly, into 80\% for training and 20\% for testing, e.g., for 50 Salads with 50 videos in total, we use 40 videos for training and 10 videos for testing. Tab.~\ref{tab:generalization_results} presents the results of our method and CTE~\cite{kukleva2019unsupervised}. As expected, the results of all methods decline as compared to those reported in preceding sections. In addition, our method  continues to outperform CTE in this experiment setup. 

\begin{table}[t]
\begin{minipage}{1.0\linewidth}
\captionsetup{type=table}
\centering
{
\begin{tabular}{l|l|l}
\specialrule{1pt}{1pt}{1pt}
\textbf{Approach} &  \textbf{F1-Score} & \textbf{MOF}\\

\midrule
CTE~\cite{kukleva2019unsupervised} & 44.9 & 47.6 \\

\rowcolor{beaublue} Ours (TOT) &  \textit{\underline{51.7}} & \textit{\underline{56.3}}  \\

\rowcolor{beaublue} Ours (TOT+TCL) & \textbf{53.4}    & \textbf{58.1}  \\

\specialrule{1pt}{1pt}{1pt}
\end{tabular}
}
\caption{Results on Desktop Assembly. The best results are in \textbf{bold}. The second best are \textit{\underline{underlined}}.}
\label{tab:da_results}
\end{minipage}

\begin{minipage}{1.0\linewidth}
\centering
\captionsetup{type=table}
\begin{tabular}{l|l|l|l}
\specialrule{1pt}{1pt}{1pt} 
\textbf{Dataset} & \textbf{Approach} &  \textbf{F1-Score} & \textbf{MOF}\\

\midrule
\multirow{3}{*}{\textbf{E}}

& CTE~\cite{kukleva2019unsupervised} & 18.4 & 12.2 \\

&  \cellcolor{beaublue} Ours (TOT) &  \cellcolor{beaublue} \textit{\underline{38.2}} & \cellcolor{beaublue}  \textit{\underline{38.3}} \\

& \cellcolor{beaublue} Ours (TOT+TCL) & \cellcolor{beaublue} \textbf{44.2} & \cellcolor{beaublue} \textbf{38.6} \\

\midrule
\multirow{3}{*}{\textbf{Y}}

& CTE~\cite{kukleva2019unsupervised} & 16.4 & 17.0 \\

& \cellcolor{beaublue} Ours (TOT) & \cellcolor{beaublue}\textit{\underline{20.6}} & \cellcolor{beaublue} \textit{\underline{24.7}} \\

& \cellcolor{beaublue} Ours (TOT+TCL) & \cellcolor{beaublue} \textbf{23.6} & \cellcolor{beaublue} \textbf{38.8} \\

\midrule
\multirow{3}{*}{\textbf{B}}

& CTE~\cite{kukleva2019unsupervised} & 23.4 & \textit{\underline{40.6}} \\

& \cellcolor{beaublue} Ours (TOT) & \cellcolor{beaublue} \textit{\underline{24.5}} &  \cellcolor{beaublue} \textbf{45.3} \\

& \cellcolor{beaublue} Ours (TOT+TCL) & \cellcolor{beaublue} \textbf{25.1} & \cellcolor{beaublue} 36.1 \\

\midrule
\multirow{3}{*}{\textbf{D}}

& CTE~\cite{kukleva2019unsupervised} & 33.8 & 36.0 \\

& \cellcolor{beaublue} Ours (TOT) & \cellcolor{beaublue}  \textit{\underline{45.1}} & \cellcolor{beaublue} \textit{\underline{49.7}} \\

& \cellcolor{beaublue} Ours (TOT+TCL) & \cellcolor{beaublue} \textbf{45.4} & \cellcolor{beaublue} \textbf{51.0} \\

\specialrule{1pt}{1pt}{1pt}
\end{tabular}
\caption{Generalization results. The best results are in \textbf{bold}. The second best are \textit{\underline{underlined}}. \textbf{E} denotes 50 Salads (\emph{Eval} granularity), \textbf{Y} denotes YouTube Instructions, \textbf{B} denotes Breakfast, and \textbf{D} denotes Desktop Assembly.}
\label{tab:generalization_results}
\end{minipage}

\end{table}
\section{Conclusion}
\label{sec:conclusion}

We propose a novel approach for unsupervised activity segmentation, which jointly performs representation learning and online clustering. We introduce temporal optimal transport, which maintains the temporal order of the activity when computing pseudo-label cluster assignments. Our approach is online, processing one mini-batch at a time. We show comparable or superior performance against the state of the art on three public datasets, i.e., 50 Salads, YouTube Instructions, and Breakfast, and our Desktop Assembly dataset, while having substantially less memory requirements. One venue for our future work is to handle order variations and background frames such as VAVA~\cite{liu2021learning}. Also, our approach can be extended to include additional self-supervised losses such as visual cues~\cite{vidalmata2021joint} and action-level cues~\cite{li2021action}.
Lastly, we can utilize deep supervision~\cite{li2017deep,li2018deep,fathy2018hierarchical,zhuang2019degeneracy} for hierarchical segmentation.

\appendix
\section{Supplementary Material}

In this supplementary material, we first discuss the limitations of our method in Sec.~\ref{sec:limitation} and show some qualitative results in Sec.~\ref{sec:qualitative}. Next, we provide the details of our implementation and our Desktop Assembly dataset in Secs.~\ref{sec:implementation} and~\ref{sec:dataset} respectively. Lastly, we discuss the societal impacts of our work in Sec.~\ref{sec:societal}.

\subsection{Limitation Discussions}
\label{sec:limitation}

Below we discuss the limitations of our method, including the equal partition constraint in Eq. 6 of the main text, the fixed order prior in Eq. 8 of the main paper, the performance of TCL, the comparison with ASAL, and the case of unknown activity class.

\noindent \textbf{Equal Partition Constraint.} We impose the equal partition constraint on cluster assignments, which may not hold true for the data in practice, i.e., one action might be longer than others in a given video. However, the equal partition constraint is imposed on soft cluster assignments (cluster assignment probabilities), i.e., the sum of soft cluster assignments should be equal for all clusters. More importantly, we apply the constraint at the mini-batch level (not the dataset level), which provides some flexibility to our approach, i.e., the sum of soft cluster assignments may be equal at the mini-batch level but the final cluster assignments may favor one cluster over others to some extent. For example, it may appear in Fig. 3(c) of the main paper that the soft cluster assignments are evenly distributed, but if we obtain the hard cluster assignments (by taking max over all soft cluster assignments for each frame), we observe that cluster \#11 gets a slightly higher number of frames assigned than others. The above observations show that our approach may be capable of handling actions with various lengths to some extent, which is likely the case for the datasets used in this paper.

\noindent \textbf{Fixed Order Prior.} We apply a fixed order prior on the clusters learned via our approach. The fixed order prior allows us to introduce the temporal order-preserving constraint within the standard optimal transport module, and predict temporally ordered clusters which are more natural for video data and can be fed directly to the Viterbi decoding module at test time. As evident in Fig. 3 of the main text, OT without the fixed order prior fails to extract any temporal structure of the activity (see Fig. 3(a)), while TOT with the fixed order prior is able to capture the temporal order of the activity relatively well (see Fig. 3(c)), i.e., initial frames are assigned to cluster \#1, following frames are assigned to cluster \#2, subsequent frames are assigned to cluster \#3, and so on. The ablation study results in Tabs. 1 and 2 of the main paper show that the fixed order prior provides performance gains on 50 Salads and YouTube Instructions, which further confirms the benefits of the fixed order prior. For the datasets used in this paper, permutation generally occurs when an action is not performed by the actor. In such cases, our method assigns only a few frames to the missing action (e.g., see the yellow segment in the TOT result in Fig. 4 of the main text) and hence manages to perform relatively well on the datasets used in this work. Nevertheless, we note that if there are several permutations or missing actions, our approach may not work.

\noindent \textbf{TCL Performance.}
TCL has been used in many previous works, e.g., \cite{hadsell2006dimensionality,mobahi2009deep,goroshin2015unsupervised}, to exploit temporal cues in videos for representation learning. Specifically, it encourages neighboring video frames to be mapped to nearby points in the embedding space (or belong to the same class) and distant video frames to be mapped to far away points in the embedding space (or belong to different classes). From our experiments above, TCL works well in cases of small/medium intra-class variations, e.g., 50 Salads (\emph{Mid} granularity), YTI, and Desktop Assembly datasets, while often not performing well in cases of large intra-class variations, e.g., 50 Salads (\emph{Eval} granularity) and Breakfast datasets. Furthermore, our basic method (i.e., TOT) is able to achieve similar or better results than many previous methods on all datasets.

\noindent \textbf{ASAL Comparison.}
On the Breakfast dataset, ASAL~\cite{li2021action} performs the best, while our method (i.e., TOT) outperforms Mallow~\cite{sener2018unsupervised} and CTE\cite{kukleva2019unsupervised} and performs on par with VTE~\cite{vidalmata2021joint} and UDE~\cite{swetha2021unsupervised}. We note that ASAL is first initialized by CTE and then exploits action-level cues for refining the results of CTE (see Fig. 1 of the ASAL paper). Thus, we could instead utilize our method to provide a better initialization for ASAL and then leverage action-level cues with ASAL for boosting our performance. This remains an interesting direction for our future work. Furthermore, our method relies on a single two-layer MLP network (same as CTE), whereas ASAL employs a combination of three networks, i.e., two MLP networks and one RNN network. Since the objective of our work is to demonstrate the merit of an online clustering approach, we decide to use a single simple MLP network to facilitate a fair comparison with CTE (an offline clustering method).

\noindent \textbf{Unknown Activity Class.}
Prior works and ours assume known activity classes and known number of actions per activity. To mitigate that, in Sec.~4.7 of CTE, it proposes to make \emph{guesses} on values of $K'$ (number of activity classes) and $K$ (\emph{same} number of actions per activity), and perform multi-level clustering to predict activity classes. However, the guesses are in fact very close to the ground truth ($K'*K = 50$ vs. ground truth $48$). Our approach could be extended to perform multi-level clustering, but it is not trivial and remains our future work.

\begin{figure}[H]
     \centering
     \begin{subfigure}[ht!]{0.49\textwidth}
         \centering
         \includegraphics[width=\textwidth, trim = 0mm 5mm 10mm 0mm]{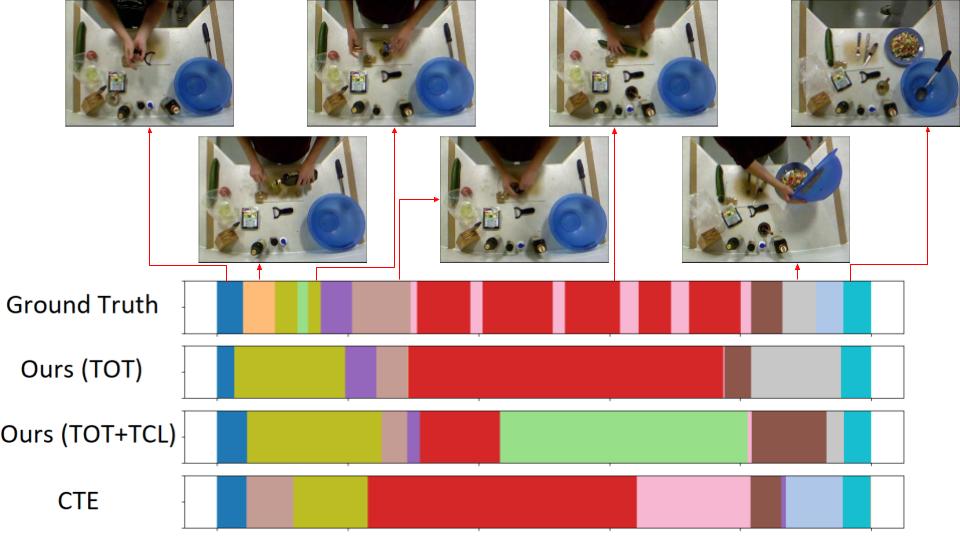}
         \caption{50 Salads (\emph{rgb-03-2}).}
         \label{fig:add_qual_fs}
     \end{subfigure}
     \hfill
     \centering
     \begin{subfigure}[ht!]{0.49\textwidth}
         \centering
         \includegraphics[width=\textwidth, trim = 0mm 5mm 10mm 0mm]{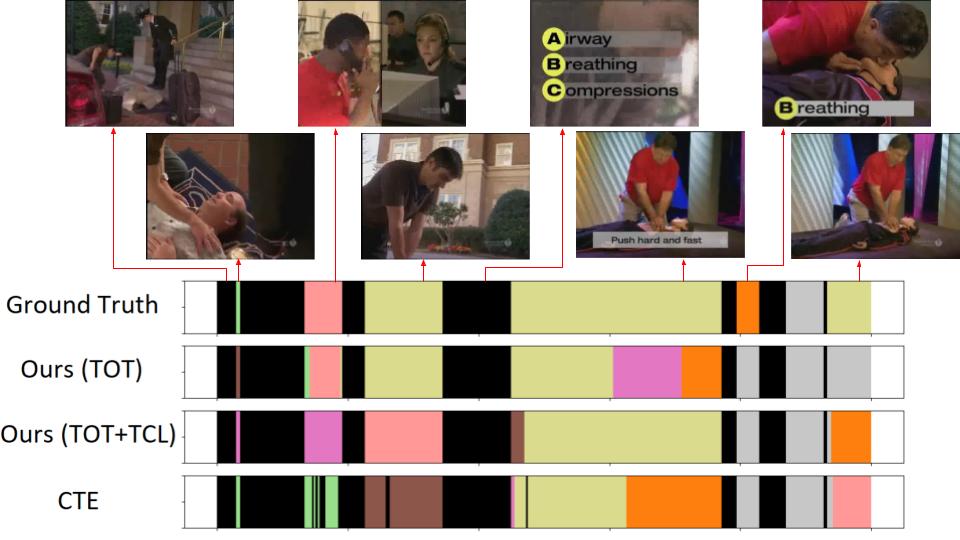}
         \caption{YouTube Instructions (\emph{cpr\_0010}).}
         \label{fig:add_qual_yti}
     \end{subfigure}
     \hfill
     \centering
     \begin{subfigure}[ht!]{0.49\textwidth}
         \centering
         \includegraphics[width=\textwidth, trim = 0mm 5mm 10mm 0mm]{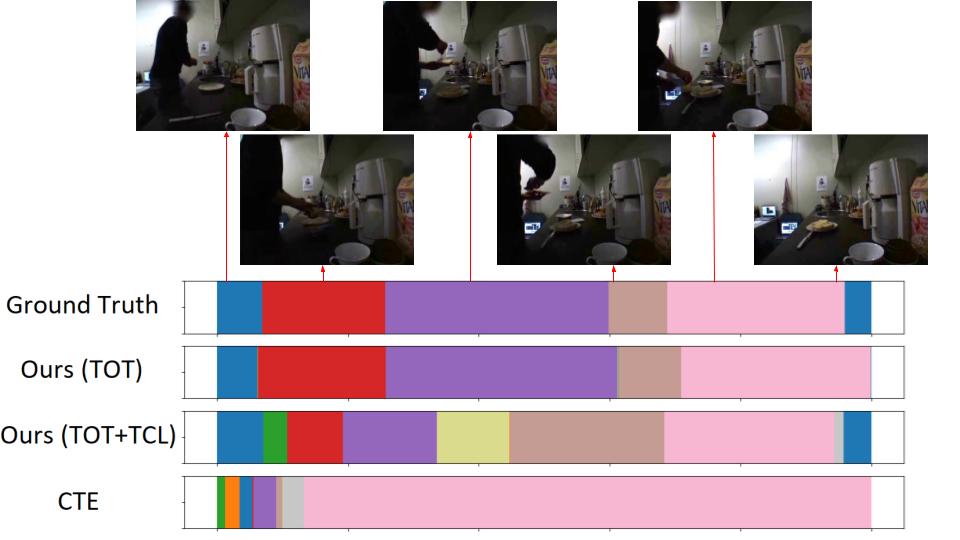}
         \caption{Breakfast (\emph{P30\_cam02\_P30\_sandwich}).}
         \label{fig:add_qual_bf}
     \end{subfigure}
     \hfill
     \centering
     \begin{subfigure}[ht!]{0.49\textwidth}
         \centering
         \includegraphics[width=\textwidth, trim = 0mm 5mm 10mm 0mm]{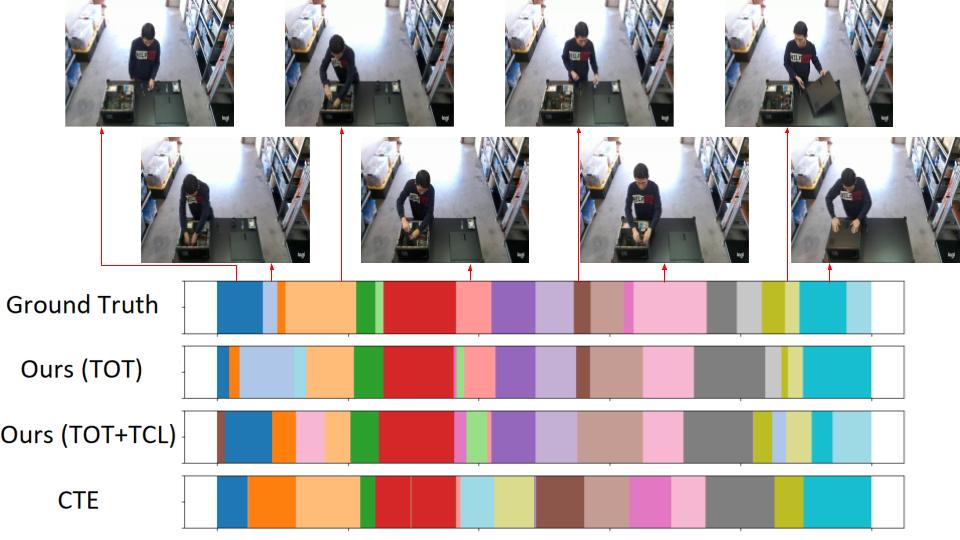}
         \caption{Desktop Assembly (\emph{2020-04-19\_13-58-20}).}
         \label{fig:add_qual_da}
     \end{subfigure}
     \caption{Qualitative Segmentation Results on all the datasets.}
     \label{fig:qual_all}
\end{figure}

\subsection{Qualitative Results}
\label{sec:qualitative}

Fig.~\ref{fig:qual_all} shows some qualitative results on 50 Salads, YouTube Instructions, Breakfast, and Desktop Assembly datasets. Overall, the results of TOT and TOT+TCL are closer to the ground truth than those of CTE~\cite{kukleva2019unsupervised}.

\subsection{Implementation Details}
\label{sec:implementation}

\noindent\textbf{Encoder Network.} As mentioned in Sec. 4 of the main paper, we employ a two-layer fully-connected encoder network on top of the pre-computed features. Each fully-connected layer is followed by the \emph{sigmoid} activation function. The dimensions of the output features are $30$, $40$ and $200$ respectively for 50 Salads, Breakfast, and YouTube Instructions datasets.

\noindent\textbf{Frame Sampling.} As we mention in Sec. 3.2 of the main text, our temporal optimal transport module assumes a fixed order of the prototypes, and assigns early frames to early prototypes and later frames to later prototypes. To implement the above, we sample frames from a video such that i) the sampled frames are temporally ordered and ii) the sampled frames spread over the entire video duration. In particular, we first divide the video into $N$ bins of equal lengths. We then sample one anchor frame $\boldsymbol{z}_i$ from the $i$-th bin with $i \in \{1, 2, ..., N\}$. For the temporal coherence loss presented in Sec. 3.1 of the main paper, we sample a ``positive'' frame $\boldsymbol{z}^+_i$ for each anchor frame $\boldsymbol{z}_i$, i.e., $\boldsymbol{z}^+_i$ is inside a temporal window of $\lambda$ from $\boldsymbol{z}_i$. Further, we consider all $\boldsymbol{z}^+_j$ with $j\neq i$ as ``negative'' frames for $\boldsymbol{z}_i$.

\noindent\textbf{Background Class on Breakfast.} 
The ``SIL'' action class in the Breakfast dataset corresponds to both background frames occurring at the start and at the end of the videos. However, the background frames at the start of the videos are visually and temporally different from those at the end of the videos. Therefore, following the 50 Salads dataset, we break the starting background frames and the ending background frames into 2 separate action classes (i.e., ``action\_start'' and ``action\_end''). For a fair comparison, we have also evaluated CTE~\cite{kukleva2019unsupervised} with the above background label splitting, however that leads to performance drops on both F1-Score and MOF metrics. In particular, CTE with background label splitting obtains $22.7$ F1-Score and $41.5\%$ MOF, whereas CTE without background label splitting (in Tab. 5 of the main text) achieves $26.4$ F1-Score and $41.8\%$ MOF.

\noindent\textbf{Adding Entropy Regularization to Eq.~9.} 
The entropy regularization in Eq.~5 ensures cluster assignments are smoothly spread out among clusters but does not consider temporal positions of frames. The KL divergence in Eq.~9 takes both factors into account by considering temporal positions of frames and imposing a smooth prior distribution (Eq.~8) on cluster assignments. We did try adding the entropy regularization to Eq.~9 but did not get better results (for TOT on 50 Salads - \emph{Eval} granularity, we obtained $46.2$\% vs. $47.4$\% in Tab.~3). Thus, we did not include the entropy regularization term in Eq.~9.

\noindent\textbf{Hyperparameter Settings.} The network is trained by using the ADAM optimizer~\cite{kingma2014adam} at a learning rate of $10^{-3}$ and a weight decay of $10^{-4}$. We freeze the gradients for the prototypes during the first few iterations for better convergence~\cite{caron2020unsupervised}. For the three public datasets, we set $\tau$ to 0.1, $\lambda$ to $30$, and $\alpha$ to $1.0$. Further, the number of Sinkhorn-Knopp iterations is fixed to $3$ and each mini-batch contains sampled frames from $2$ videos. Tabs.~\ref{tab:hyperparameters_tot} and~\ref{tab:hyperparameters_tot_tcn} present the hyperparameter settings for TOT and TOT+TCL respectively on the three public datasets, including 50 Salads, YouTube Instructions, and Breakfast.

\noindent\textbf{Computing Resources.} Our experiments are conducted with a single Nvidia V100 GPU on Microsoft Azure. 

\begin{table*}[ht!]
\begin{minipage}[t]{1.0\linewidth}
\centering
{
\begin{tabular}{l|l}

\specialrule{1pt}{1pt}{1pt}

\textbf{Hyperparameter} &  \textbf{Value}\\

\midrule
Rho ($\rho$) &  $0.07$ (\textbf{E}), $0.08$ (\textbf{M}), $0.08$ (\textbf{Y}), $0.05$ (\textbf{B})   \\

Sigma ($\sigma$) &   $2.5$ (\textbf{E}), $2.0$ (\textbf{M}), $1.25$ (\textbf{Y}), $1.0$ (\textbf{B})  \\

Mini-batch size &  $512$ \\

Temperature ($\tau$) & $0.1$ \\

Number of Sinkhorn-Knopp iterations & $3$\\

Learning rate  & $10^{-3}$\\

Weight decay & $10^{-4}$\\

Number of videos per mini-batch & $2$ \\

\specialrule{1pt}{1pt}{1pt}
\end{tabular}
}
\caption{Hyperparameter settings for TOT on the three public datasets, including 50 Salads, YouTube Instructions, and Breakfast. \textbf{E} denotes 50 Salads (\emph{Eval} granularity), \textbf{M} denotes 50 Salads (\emph{Mid} granularity), \textbf{Y} denotes YouTube Instructions, and \textbf{B} denotes Breakfast.}
\label{tab:hyperparameters_tot}

\end{minipage}

\end{table*}
\begin{table*}[ht!]
\begin{minipage}[t]{1.0\linewidth}
\centering
{
\begin{tabular}{l|l}

\specialrule{1pt}{1pt}{1pt}

\textbf{Hyperparameter} &  \textbf{Value}\\

\midrule
Rho ($\rho$) &  $0.08$ (\textbf{E}), $0.07$ (\textbf{M}), $0.07$ (\textbf{Y}), $0.04$ (\textbf{B})   \\

Sigma ($\sigma$) &   $2.5$ (\textbf{E}), $1.75$ (\textbf{M}), $3.0$ (\textbf{Y}), $0.75$ (\textbf{B})  \\

Mini-batch size &  $512$ \\

Temperature ($\tau$) & $0.1$ \\

Window size ($\lambda$) & $30$ \\

Alpha ($\alpha$) & $1.0$ \\

Number of Sinkhorn-Knopp iterations & $3$\\

Learning rate  & $10^{-3}$\\

Weight decay & $10^{-4}$\\

Number of videos per mini-batch & $2$ \\

\specialrule{1pt}{1pt}{1pt}
\end{tabular}
}
\caption{Hyperparameter settings for TOT+TCL on the three public datasets, including  50 Salads, YouTube Instructions, and Breakfast. \textbf{E} denotes 50 Salads (\emph{Eval} granularity), \textbf{M} denotes 50 Salads (\emph{Mid} granularity), \textbf{Y} denotes YouTube Instructions, and \textbf{B} denotes Breakfast.}
\label{tab:hyperparameters_tot_tcn}

\end{minipage}

\end{table*}

\subsection{Desktop Assembly Dataset Details}
\label{sec:dataset}

Our Desktop Assembly dataset includes $76$ videos of different actors assembling a desktop computer from its parts. The desktop assembly activity consists of $22$ action classes and $1$ background class, amounting to a total of $23$ action classes. The actions are ``picking up chip'', ``placing chip on motherboard'', ``closing cover'', ``picking up screw and screw driver'', ``tightening screw'', ``plugging stick in'', ``picking up fan'', ``placing fan on motherboard'', ``tightening screw A'', ``tightening screw B'', ``tightening screw C'', ``tightening screw D'', ``putting screw driver down'', ``connecting wire to motherboard'', ``picking up RAM'', ``installing RAM'', ``locking RAM'', ``picking up disk'', ``installing disk'', ``connecting wire A to motherboard'', ``connecting wire B to motherboard'', ``closing lid'', and ``background''. The activity is performed by $4$ different actors with various appearances, speeds, and viewpoints. We downsample the videos to $10$ frames per second, resulting in a total of $59,165$ frames for the entire dataset. We use ResNet-18~\cite{he2016deep} pre-trained on ImageNet to obtain pre-computed features which are used as input for all methods. The original videos, pre-computed features, and action class labels are available at \url{https://bit.ly/3JKm0JP}. We note that the action class labels are only used during evaluation. Our hyperparameter settings for TOT and TOT+TCL on our Desktop Assembly dataset are presented in Tab.~\ref{tab:da_hyperparameters}.

\subsection{Societal Impacts}
\label{sec:societal}
Our approach enables learning video recognition models without requiring action labels. It would positively impact the problems of worker training and assistance, where models automatically built from video datasets of expert demonstrations in diverse domains, e.g., factory work and medical surgery, could be used to provide training and guidance to new workers. Similarly, there exist problems such as surgery standardization, where operation room video datasets could be processed with approaches such as ours to improve the standard of care for patients globally. On the other hand, video understanding algorithms could generally be used in surveillance applications, where they improve security and productivity at the cost of privacy.

\begin{table*}[ht!]
\begin{minipage}[t]{1.0\linewidth}
\centering
{%
\begin{tabular}{l|l}

\specialrule{1pt}{1pt}{1pt}

\textbf{Hyperparameter} &  \textbf{Value}\\
\midrule
Rho ($\rho$) & $0.07$ \\

Sigma ($\sigma$) & $2.0$ \\

Mini-batch size &  $512$ \\

Temperature ($\tau$) & $0.1$ \\

Window size ($\lambda$) & $30$ \\

Alpha ($\alpha$) & $1.0$ \\

Number of Sinkhorn-Knopp iterations & $3$\\

Learning rate  & $10^{-3}$\\

Weight decay & $10^{-4}$\\

Number of videos per mini-batch & $2$ \\

\specialrule{1pt}{1pt}{1pt}
\end{tabular}
}
\caption{Hyperparameter settings for TOT and TOT+TCL on our Desktop Assembly dataset. Window size ($\lambda$) and Alpha ($\alpha$) are only used in TOT+TCL.}
\label{tab:da_hyperparameters}
\end{minipage}
\end{table*}

{\small
\bibliographystyle{ieee_fullname}
\bibliography{references}

\begin{thebibliography}{10}\itemsep=-1pt

\bibitem{aakur2019perceptual}
Sathyanarayanan~N Aakur and Sudeep Sarkar.
\newblock A perceptual prediction framework for self supervised event
  segmentation.
\newblock In {\em Proceedings of the IEEE/CVF Conference on Computer Vision and
  Pattern Recognition}, pages 1197--1206, 2019.

\bibitem{ahsan2018discrimnet}
Unaiza Ahsan, Chen Sun, and Irfan Essa.
\newblock Discrimnet: Semi-supervised action recognition from videos using
  generative adversarial networks.
\newblock {\em arXiv preprint arXiv:1801.07230}, 2018.

\bibitem{alayrac2016unsupervised}
Jean-Baptiste Alayrac, Piotr Bojanowski, Nishant Agrawal, Josef Sivic, Ivan
  Laptev, and Simon Lacoste-Julien.
\newblock Unsupervised learning from narrated instruction videos.
\newblock In {\em Proceedings of the IEEE Conference on Computer Vision and
  Pattern Recognition}, pages 4575--4583, 2016.

\bibitem{asano2019self}
YM Asano, C Rupprecht, and A Vedaldi.
\newblock Self-labelling via simultaneous clustering and representation
  learning.
\newblock In {\em International Conference on Learning Representations}, 2019.

\bibitem{bautista2016cliquecnn}
Miguel~{\'A}ngel Bautista, Artsiom Sanakoyeu, Ekaterina Tikhoncheva, and
  Bj{\"o}rn Ommer.
\newblock Cliquecnn: Deep unsupervised exemplar learning.
\newblock In {\em NIPS}, 2016.

\bibitem{bengio2009slow}
Yoshua Bengio and James~S Bergstra.
\newblock Slow, decorrelated features for pretraining complex cell-like
  networks.
\newblock In {\em Advances in neural information processing systems}, pages
  99--107, 2009.

\bibitem{carlucci2019domain}
Fabio~M Carlucci, Antonio D'Innocente, Silvia Bucci, Barbara Caputo, and
  Tatiana Tommasi.
\newblock Domain generalization by solving jigsaw puzzles.
\newblock In {\em Proceedings of the IEEE Conference on Computer Vision and
  Pattern Recognition}, pages 2229--2238, 2019.

\bibitem{caron2018deep}
Mathilde Caron, Piotr Bojanowski, Armand Joulin, and Matthijs Douze.
\newblock Deep clustering for unsupervised learning of visual features.
\newblock In {\em Proceedings of the European Conference on Computer Vision
  (ECCV)}, pages 132--149, 2018.

\bibitem{caron2019unsupervised}
Mathilde Caron, Piotr Bojanowski, Julien Mairal, and Armand Joulin.
\newblock Unsupervised pre-training of image features on non-curated data.
\newblock In {\em Proceedings of the IEEE/CVF International Conference on
  Computer Vision}, pages 2959--2968, 2019.

\bibitem{caron2020unsupervised}
Mathilde Caron, Ishan Misra, Julien Mairal, Priya Goyal, Piotr Bojanowski, and
  Armand Joulin.
\newblock Unsupervised learning of visual features by contrasting cluster
  assignments.
\newblock In {\em Neural Information Processing Systems}, 2020.

\bibitem{carreira2017quo}
Joao Carreira and Andrew Zisserman.
\newblock Quo vadis, action recognition? a new model and the kinetics dataset.
\newblock In {\em proceedings of the IEEE Conference on Computer Vision and
  Pattern Recognition}, pages 6299--6308, 2017.

\bibitem{chang2019d3tw}
Chien-Yi Chang, De-An Huang, Yanan Sui, Li Fei-Fei, and Juan~Carlos Niebles.
\newblock D3tw: Discriminative differentiable dynamic time warping for weakly
  supervised action alignment and segmentation.
\newblock In {\em Proceedings of the IEEE/CVF Conference on Computer Vision and
  Pattern Recognition}, pages 3546--3555, 2019.

\bibitem{chao2018rethinking}
Yu-Wei Chao, Sudheendra Vijayanarasimhan, Bryan Seybold, David~A Ross, Jia
  Deng, and Rahul Sukthankar.
\newblock Rethinking the faster r-cnn architecture for temporal action
  localization.
\newblock In {\em Proceedings of the IEEE Conference on Computer Vision and
  Pattern Recognition}, pages 1130--1139, 2018.

\bibitem{chen2020action}
Min-Hung Chen, Baopu Li, Yingze Bao, Ghassan AlRegib, and Zsolt Kira.
\newblock Action segmentation with joint self-supervised temporal domain
  adaptation.
\newblock In {\em Proceedings of the IEEE/CVF Conference on Computer Vision and
  Pattern Recognition}, pages 9454--9463, 2020.

\bibitem{choi2020shuffle}
Jinwoo Choi, Gaurav Sharma, Samuel Schulter, and Jia-Bin Huang.
\newblock Shuffle and attend: Video domain adaptation.
\newblock In {\em European Conference on Computer Vision}, pages 678--695.
  Springer, 2020.

\bibitem{cuturi2013sinkhorn}
Marco Cuturi.
\newblock Sinkhorn distances: Lightspeed computation of optimal transport.
\newblock {\em Advances in neural information processing systems},
  26:2292--2300, 2013.

\bibitem{diba2019dynamonet}
Ali Diba, Vivek Sharma, Luc~Van Gool, and Rainer Stiefelhagen.
\newblock Dynamonet: Dynamic action and motion network.
\newblock In {\em Proceedings of the IEEE International Conference on Computer
  Vision}, pages 6192--6201, 2019.

\bibitem{ding2018weakly}
Li Ding and Chenliang Xu.
\newblock Weakly-supervised action segmentation with iterative soft boundary
  assignment.
\newblock In {\em Proceedings of the IEEE Conference on Computer Vision and
  Pattern Recognition}, pages 6508--6516, 2018.

\bibitem{dwibedi2019temporal}
Debidatta Dwibedi, Yusuf Aytar, Jonathan Tompson, Pierre Sermanet, and Andrew
  Zisserman.
\newblock Temporal cycle-consistency learning.
\newblock In {\em Proceedings of the IEEE Conference on Computer Vision and
  Pattern Recognition}, pages 1801--1810, 2019.

\bibitem{fathy2018hierarchical}
Mohammed~E Fathy, Quoc-Huy Tran, M Zeeshan~Zia, Paul Vernaza, and Manmohan
  Chandraker.
\newblock Hierarchical metric learning and matching for 2d and 3d geometric
  correspondences.
\newblock In {\em Proceedings of the European Conference on Computer Vision
  (ECCV)}, pages 803--819, 2018.

\bibitem{fayyaz2020sct}
Mohsen Fayyaz and Jurgen Gall.
\newblock Sct: Set constrained temporal transformer for set supervised action
  segmentation.
\newblock In {\em Proceedings of the IEEE/CVF Conference on Computer Vision and
  Pattern Recognition}, pages 501--510, 2020.

\bibitem{feng2019self}
Zeyu Feng, Chang Xu, and Dacheng Tao.
\newblock Self-supervised representation learning by rotation feature
  decoupling.
\newblock In {\em Proceedings of the IEEE Conference on Computer Vision and
  Pattern Recognition}, pages 10364--10374, 2019.

\bibitem{fernando2017self}
Basura Fernando, Hakan Bilen, Efstratios Gavves, and Stephen Gould.
\newblock Self-supervised video representation learning with odd-one-out
  networks.
\newblock In {\em Proceedings of the IEEE conference on computer vision and
  pattern recognition}, pages 3636--3645, 2017.

\bibitem{gammulle2019predicting}
Harshala Gammulle, Simon Denman, Sridha Sridharan, and Clinton Fookes.
\newblock Predicting the future: A jointly learnt model for action
  anticipation.
\newblock In {\em Proceedings of the IEEE International Conference on Computer
  Vision}, pages 5562--5571, 2019.

\bibitem{gidaris2020learning}
Spyros Gidaris, Andrei Bursuc, Nikos Komodakis, Patrick P{\'e}rez, and Matthieu
  Cord.
\newblock Learning representations by predicting bags of visual words.
\newblock In {\em Proceedings of the IEEE/CVF Conference on Computer Vision and
  Pattern Recognition}, pages 6928--6938, 2020.

\bibitem{gidaris2018unsupervised}
Spyros Gidaris, Praveer Singh, and Nikos Komodakis.
\newblock Unsupervised representation learning by predicting image rotations.
\newblock In {\em International Conference on Learning Representations}, 2018.

\bibitem{gong2019memorizing}
Dong Gong, Lingqiao Liu, Vuong Le, Budhaditya Saha, Moussa~Reda Mansour, Svetha
  Venkatesh, and Anton van~den Hengel.
\newblock Memorizing normality to detect anomaly: Memory-augmented deep
  autoencoder for unsupervised anomaly detection.
\newblock In {\em Proceedings of the IEEE/CVF International Conference on
  Computer Vision}, pages 1705--1714, 2019.

\bibitem{goroshin2015unsupervised}
Ross Goroshin, Joan Bruna, Jonathan Tompson, David Eigen, and Yann LeCun.
\newblock Unsupervised learning of spatiotemporally coherent metrics.
\newblock In {\em Proceedings of the IEEE international conference on computer
  vision}, pages 4086--4093, 2015.

\bibitem{hadsell2006dimensionality}
Raia Hadsell, Sumit Chopra, and Yann LeCun.
\newblock Dimensionality reduction by learning an invariant mapping.
\newblock In {\em 2006 IEEE Computer Society Conference on Computer Vision and
  Pattern Recognition (CVPR'06)}, volume~2, pages 1735--1742. IEEE, 2006.

\bibitem{han2019video}
Tengda Han, Weidi Xie, and Andrew Zisserman.
\newblock Video representation learning by dense predictive coding.
\newblock In {\em Proceedings of the IEEE International Conference on Computer
  Vision Workshops}, pages 0--0, 2019.

\bibitem{haresh2021learning}
Sanjay Haresh, Sateesh Kumar, Huseyin Coskun, Shahram~Najam Syed, Andrey Konin,
  Muhammad~Zeeshan Zia, and Quoc-Huy Tran.
\newblock Learning by aligning videos in time.
\newblock In {\em Proceedings of the IEEE/CVF Conference on Computer Vision and
  Pattern Recognition}, 2021.

\bibitem{haresh2020towards}
Sanjay Haresh, Sateesh Kumar, M~Zeeshan Zia, and Quoc-Huy Tran.
\newblock Towards anomaly detection in dashcam videos.
\newblock In {\em 2020 IEEE Intelligent Vehicles Symposium (IV)}, pages
  1407--1414. IEEE.

\bibitem{he2016deep}
Kaiming He, Xiangyu Zhang, Shaoqing Ren, and Jian Sun.
\newblock Deep residual learning for image recognition.
\newblock In {\em Proceedings of the IEEE conference on computer vision and
  pattern recognition}, pages 770--778, 2016.

\bibitem{hinton1994autoencoders}
Geoffrey~E Hinton and Richard~S Zemel.
\newblock Autoencoders, minimum description length and helmholtz free energy.
\newblock In {\em Advances in neural information processing systems}, pages
  3--10, 1994.

\bibitem{huang2016connectionist}
De-An Huang, Li Fei-Fei, and Juan~Carlos Niebles.
\newblock Connectionist temporal modeling for weakly supervised action
  labeling.
\newblock In {\em European Conference on Computer Vision}, pages 137--153.
  Springer, 2016.

\bibitem{huang2019unsupervised}
Jiabo Huang, Qi Dong, Shaogang Gong, and Xiatian Zhu.
\newblock Unsupervised deep learning by neighbourhood discovery.
\newblock In {\em International Conference on Machine Learning}, pages
  2849--2858. PMLR, 2019.

\bibitem{kim2019self}
Dahun Kim, Donghyeon Cho, and In~So Kweon.
\newblock Self-supervised video representation learning with space-time cubic
  puzzles.
\newblock In {\em Proceedings of the AAAI Conference on Artificial
  Intelligence}, volume~33, pages 8545--8552, 2019.

\bibitem{kim2018learning}
Dahun Kim, Donghyeon Cho, Donggeun Yoo, and In~So Kweon.
\newblock Learning image representations by completing damaged jigsaw puzzles.
\newblock In {\em 2018 IEEE Winter Conference on Applications of Computer
  Vision (WACV)}, pages 793--802. IEEE, 2018.

\bibitem{kingma2014adam}
Diederik~P Kingma and Jimmy Ba.
\newblock Adam: A method for stochastic optimization.
\newblock {\em arXiv preprint arXiv:1412.6980}, 2014.

\bibitem{kuehne2014language}
Hilde Kuehne, Ali Arslan, and Thomas Serre.
\newblock The language of actions: Recovering the syntax and semantics of
  goal-directed human activities.
\newblock In {\em Proceedings of the IEEE conference on computer vision and
  pattern recognition}, pages 780--787, 2014.

\bibitem{kuehne2016end}
Hilde Kuehne, Juergen Gall, and Thomas Serre.
\newblock An end-to-end generative framework for video segmentation and
  recognition.
\newblock In {\em 2016 IEEE Winter Conference on Applications of Computer
  Vision (WACV)}, pages 1--8. IEEE, 2016.

\bibitem{kuehne2017weakly}
Hilde Kuehne, Alexander Richard, and Juergen Gall.
\newblock Weakly supervised learning of actions from transcripts.
\newblock {\em Computer Vision and Image Understanding}, 163:78--89, 2017.

\bibitem{kukleva2019unsupervised}
Anna Kukleva, Hilde Kuehne, Fadime Sener, and Jurgen Gall.
\newblock Unsupervised learning of action classes with continuous temporal
  embedding.
\newblock In {\em Proceedings of the IEEE/CVF Conference on Computer Vision and
  Pattern Recognition}, pages 12066--12074, 2019.

\bibitem{larsson2016learning}
Gustav Larsson, Michael Maire, and Gregory Shakhnarovich.
\newblock Learning representations for automatic colorization.
\newblock In {\em European conference on computer vision}, pages 577--593.
  Springer, 2016.

\bibitem{larsson2017colorization}
Gustav Larsson, Michael Maire, and Gregory Shakhnarovich.
\newblock Colorization as a proxy task for visual understanding.
\newblock In {\em Proceedings of the IEEE Conference on Computer Vision and
  Pattern Recognition}, pages 6874--6883, 2017.

\bibitem{lea2016segmental}
Colin Lea, Austin Reiter, Ren{\'e} Vidal, and Gregory~D Hager.
\newblock Segmental spatiotemporal cnns for fine-grained action segmentation.
\newblock In {\em European Conference on Computer Vision}, pages 36--52.
  Springer, 2016.

\bibitem{lee2017unsupervised}
Hsin-Ying Lee, Jia-Bin Huang, Maneesh Singh, and Ming-Hsuan Yang.
\newblock Unsupervised representation learning by sorting sequences.
\newblock In {\em Proceedings of the IEEE International Conference on Computer
  Vision}, pages 667--676, 2017.

\bibitem{li2018deep}
C. Li, M.~Z. Zia, Q. Tran, X. Yu, G.~D. Hager, and M. Chandraker.
\newblock Deep supervision with intermediate concepts.
\newblock {\em IEEE Transactions on Pattern Analysis and Machine Intelligence},
  pages 1--1, 2018.

\bibitem{li2017deep}
Chi Li, M~Zeeshan Zia, Quoc-Huy Tran, Xiang Yu, Gregory~D Hager, and Manmohan
  Chandraker.
\newblock Deep supervision with shape concepts for occlusion-aware 3d object
  parsing.
\newblock In {\em 2017 IEEE Conference on Computer Vision and Pattern
  Recognition (CVPR)}, pages 388--397. IEEE, 2017.

\bibitem{li2019weakly}
Jun Li, Peng Lei, and Sinisa Todorovic.
\newblock Weakly supervised energy-based learning for action segmentation.
\newblock In {\em Proceedings of the IEEE/CVF International Conference on
  Computer Vision}, pages 6243--6251, 2019.

\bibitem{li2020set}
Jun Li and Sinisa Todorovic.
\newblock Set-constrained viterbi for set-supervised action segmentation.
\newblock In {\em Proceedings of the IEEE/CVF Conference on Computer Vision and
  Pattern Recognition}, pages 10820--10829, 2020.

\bibitem{li2021action}
Jun Li and Sinisa Todorovic.
\newblock Action shuffle alternating learning for unsupervised action
  segmentation.
\newblock In {\em Proceedings of the IEEE/CVF Conference on Computer Vision and
  Pattern Recognition}, 2021.

\bibitem{li2020ms}
Shi-Jie Li, Yazan AbuFarha, Yun Liu, Ming-Ming Cheng, and Juergen Gall.
\newblock Ms-tcn++: Multi-stage temporal convolutional network for action
  segmentation.
\newblock {\em IEEE Transactions on Pattern Analysis and Machine Intelligence},
  2020.

\bibitem{li2021temporal}
Zhe Li, Yazan Abu~Farha, and Jurgen Gall.
\newblock Temporal action segmentation from timestamp supervision.
\newblock In {\em Proceedings of the IEEE/CVF Conference on Computer Vision and
  Pattern Recognition}, pages 8365--8374, 2021.

\bibitem{liu2021learning}
Weizhe Liu, Bugra Tekin, Huseyin Coskun, Vibhav Vineet, Pascal Fua, and Marc
  Pollefeys.
\newblock Learning to align sequential actions in the wild.
\newblock In {\em Proceedings of the IEEE/CVF Conference on Computer Vision and
  Pattern Recognition}, 2022.

\bibitem{liu2018leveraging}
Xialei Liu, Joost Van De~Weijer, and Andrew~D Bagdanov.
\newblock Leveraging unlabeled data for crowd counting by learning to rank.
\newblock In {\em Proceedings of the IEEE Conference on Computer Vision and
  Pattern Recognition}, pages 7661--7669, 2018.

\bibitem{malmaud2015s}
Jonathan Malmaud, Jonathan Huang, Vivek Rathod, Nicholas Johnston, Andrew
  Rabinovich, and Kevin Murphy.
\newblock What's cookin'? interpreting cooking videos using text, speech and
  vision.
\newblock In {\em HLT-NAACL}, 2015.

\bibitem{misra2016shuffle}
Ishan Misra, C~Lawrence Zitnick, and Martial Hebert.
\newblock Shuffle and learn: unsupervised learning using temporal order
  verification.
\newblock In {\em European Conference on Computer Vision}, pages 527--544.
  Springer, 2016.

\bibitem{mobahi2009deep}
Hossein Mobahi, Ronan Collobert, and Jason Weston.
\newblock Deep learning from temporal coherence in video.
\newblock In {\em Proceedings of the 26th Annual International Conference on
  Machine Learning}, pages 737--744, 2009.

\bibitem{noroozi2017representation}
Mehdi Noroozi, Hamed Pirsiavash, and Paolo Favaro.
\newblock Representation learning by learning to count.
\newblock In {\em Proceedings of the IEEE International Conference on Computer
  Vision}, pages 5898--5906, 2017.

\bibitem{richard2016temporal}
Alexander Richard and Juergen Gall.
\newblock Temporal action detection using a statistical language model.
\newblock In {\em Proceedings of the IEEE Conference on Computer Vision and
  Pattern Recognition}, pages 3131--3140, 2016.

\bibitem{richard2017weakly}
Alexander Richard, Hilde Kuehne, and Juergen Gall.
\newblock Weakly supervised action learning with rnn based fine-to-coarse
  modeling.
\newblock In {\em Proceedings of the IEEE Conference on Computer Vision and
  Pattern Recognition}, pages 754--763, 2017.

\bibitem{richard2018action}
Alexander Richard, Hilde Kuehne, and Juergen Gall.
\newblock Action sets: Weakly supervised action segmentation without ordering
  constraints.
\newblock In {\em Proceedings of the IEEE conference on Computer Vision and
  Pattern Recognition}, pages 5987--5996, 2018.

\bibitem{richard2018neuralnetwork}
Alexander Richard, Hilde Kuehne, Ahsan Iqbal, and Juergen Gall.
\newblock Neuralnetwork-viterbi: A framework for weakly supervised video
  learning.
\newblock In {\em Proceedings of the IEEE Conference on Computer Vision and
  Pattern Recognition}, pages 7386--7395, 2018.

\bibitem{sener2018unsupervised}
Fadime Sener and Angela Yao.
\newblock Unsupervised learning and segmentation of complex activities from
  video.
\newblock In {\em Proceedings of the IEEE Conference on Computer Vision and
  Pattern Recognition}, pages 8368--8376, 2018.

\bibitem{sener2015unsupervised}
Ozan Sener, Amir~R Zamir, Silvio Savarese, and Ashutosh Saxena.
\newblock Unsupervised semantic parsing of video collections.
\newblock In {\em Proceedings of the IEEE International Conference on Computer
  Vision}, pages 4480--4488, 2015.

\bibitem{sermanet2018time}
Pierre Sermanet, Corey Lynch, Yevgen Chebotar, Jasmine Hsu, Eric Jang, Stefan
  Schaal, Sergey Levine, and Google Brain.
\newblock Time-contrastive networks: Self-supervised learning from video.
\newblock In {\em 2018 IEEE International Conference on Robotics and Automation
  (ICRA)}, pages 1134--1141. IEEE, 2018.

\bibitem{shou2017cdc}
Zheng Shou, Jonathan Chan, Alireza Zareian, Kazuyuki Miyazawa, and Shih-Fu
  Chang.
\newblock Cdc: Convolutional-de-convolutional networks for precise temporal
  action localization in untrimmed videos.
\newblock In {\em Proceedings of the IEEE conference on computer vision and
  pattern recognition}, pages 5734--5743, 2017.

\bibitem{shou2016temporal}
Zheng Shou, Dongang Wang, and Shih-Fu Chang.
\newblock Temporal action localization in untrimmed videos via multi-stage
  cnns.
\newblock In {\em Proceedings of the IEEE conference on computer vision and
  pattern recognition}, pages 1049--1058, 2016.

\bibitem{sohn2016improved}
Kihyuk Sohn.
\newblock Improved deep metric learning with multi-class n-pair loss objective.
\newblock In {\em Proceedings of the 30th International Conference on Neural
  Information Processing Systems}, pages 1857--1865, 2016.

\bibitem{srivastava2015unsupervised}
Nitish Srivastava, Elman Mansimov, and Ruslan Salakhudinov.
\newblock Unsupervised learning of video representations using lstms.
\newblock In {\em International conference on machine learning}, pages
  843--852, 2015.

\bibitem{stein2013combining}
Sebastian Stein and Stephen~J McKenna.
\newblock Combining embedded accelerometers with computer vision for
  recognizing food preparation activities.
\newblock In {\em Proceedings of the 2013 ACM international joint conference on
  Pervasive and ubiquitous computing}, pages 729--738, 2013.

\bibitem{su2017order}
Bing Su and Gang Hua.
\newblock Order-preserving wasserstein distance for sequence matching.
\newblock In {\em Proceedings of the IEEE conference on computer vision and
  pattern recognition}, pages 1049--1057, 2017.

\bibitem{sultani2018real}
Waqas Sultani, Chen Chen, and Mubarak Shah.
\newblock Real-world anomaly detection in surveillance videos.
\newblock In {\em Proceedings of the IEEE conference on computer vision and
  pattern recognition}, pages 6479--6488, 2018.

\bibitem{swetha2021unsupervised}
Sirnam Swetha, Hilde Kuehne, Yogesh~S Rawat, and Mubarak Shah.
\newblock Unsupervised discriminative embedding for sub-action learning in
  complex activities.
\newblock In {\em 2021 IEEE International Conference on Image Processing
  (ICIP)}, pages 2588--2592. IEEE, 2021.

\bibitem{tran2015learning}
Du Tran, Lubomir Bourdev, Rob Fergus, Lorenzo Torresani, and Manohar Paluri.
\newblock Learning spatiotemporal features with 3d convolutional networks.
\newblock In {\em Proceedings of the IEEE international conference on computer
  vision}, pages 4489--4497, 2015.

\bibitem{tran2018closer}
Du Tran, Heng Wang, Lorenzo Torresani, Jamie Ray, Yann LeCun, and Manohar
  Paluri.
\newblock A closer look at spatiotemporal convolutions for action recognition.
\newblock In {\em Proceedings of the IEEE conference on Computer Vision and
  Pattern Recognition}, pages 6450--6459, 2018.

\bibitem{vidalmata2021joint}
Rosaura~G VidalMata, Walter~J Scheirer, Anna Kukleva, David Cox, and Hilde
  Kuehne.
\newblock Joint visual-temporal embedding for unsupervised learning of actions
  in untrimmed sequences.
\newblock In {\em Proceedings of the IEEE/CVF Winter Conference on Applications
  of Computer Vision}, pages 1238--1247, 2021.

\bibitem{vincent2008extracting}
Pascal Vincent, Hugo Larochelle, Yoshua Bengio, and Pierre-Antoine Manzagol.
\newblock Extracting and composing robust features with denoising autoencoders.
\newblock In {\em Proceedings of the 25th international conference on Machine
  learning}, pages 1096--1103, 2008.

\bibitem{vondrick2016generating}
Carl Vondrick, Hamed Pirsiavash, and Antonio Torralba.
\newblock Generating videos with scene dynamics.
\newblock In {\em Advances in neural information processing systems}, pages
  613--621, 2016.

\bibitem{wang2013action}
Heng Wang and Cordelia Schmid.
\newblock Action recognition with improved trajectories.
\newblock In {\em Proceedings of the IEEE international conference on computer
  vision}, pages 3551--3558, 2013.

\bibitem{wang2018non}
Xiaolong Wang, Ross Girshick, Abhinav Gupta, and Kaiming He.
\newblock Non-local neural networks.
\newblock In {\em Proceedings of the IEEE conference on computer vision and
  pattern recognition}, pages 7794--7803, 2018.

\bibitem{wu2018unsupervised}
Zhirong Wu, Yuanjun Xiong, Stella~X Yu, and Dahua Lin.
\newblock Unsupervised feature learning via non-parametric instance
  discrimination.
\newblock In {\em Proceedings of the IEEE Conference on Computer Vision and
  Pattern Recognition}, pages 3733--3742, 2018.

\bibitem{xie2016unsupervised}
Junyuan Xie, Ross Girshick, and Ali Farhadi.
\newblock Unsupervised deep embedding for clustering analysis.
\newblock In {\em International conference on machine learning}, pages
  478--487. PMLR, 2016.

\bibitem{xu2019self}
Dejing Xu, Jun Xiao, Zhou Zhao, Jian Shao, Di Xie, and Yueting Zhuang.
\newblock Self-supervised spatiotemporal learning via video clip order
  prediction.
\newblock In {\em Proceedings of the IEEE Conference on Computer Vision and
  Pattern Recognition}, pages 10334--10343, 2019.

\bibitem{yan2020clusterfit}
Xueting Yan, Ishan Misra, Abhinav Gupta, Deepti Ghadiyaram, and Dhruv Mahajan.
\newblock Clusterfit: Improving generalization of visual representations.
\newblock In {\em Proceedings of the IEEE/CVF Conference on Computer Vision and
  Pattern Recognition}, pages 6509--6518, 2020.

\bibitem{yang2016joint}
Jianwei Yang, Devi Parikh, and Dhruv Batra.
\newblock Joint unsupervised learning of deep representations and image
  clusters.
\newblock In {\em Proceedings of the IEEE conference on computer vision and
  pattern recognition}, pages 5147--5156, 2016.

\bibitem{zeng2019graph}
Runhao Zeng, Wenbing Huang, Mingkui Tan, Yu Rong, Peilin Zhao, Junzhou Huang,
  and Chuang Gan.
\newblock Graph convolutional networks for temporal action localization.
\newblock In {\em Proceedings of the IEEE/CVF International Conference on
  Computer Vision}, pages 7094--7103, 2019.

\bibitem{zhuang2019degeneracy}
Bingbing Zhuang, Quoc-Huy Tran, Gim~Hee Lee, Loong~Fah Cheong, and Manmohan
  Chandraker.
\newblock Degeneracy in self-calibration revisited and a deep learning solution
  for uncalibrated slam.
\newblock In {\em 2019 IEEE/RSJ International Conference on Intelligent Robots
  and Systems (IROS)}, pages 3766--3773. IEEE, 2019.

\bibitem{zhuang2019local}
Chengxu Zhuang, Alex~Lin Zhai, and Daniel Yamins.
\newblock Local aggregation for unsupervised learning of visual embeddings.
\newblock In {\em Proceedings of the IEEE/CVF International Conference on
  Computer Vision}, pages 6002--6012, 2019.

\bibitem{zou2012deep}
Will Zou, Shenghuo Zhu, Kai Yu, and Andrew~Y Ng.
\newblock Deep learning of invariant features via simulated fixations in video.
\newblock In {\em Advances in neural information processing systems}, pages
  3203--3211, 2012.

\bibitem{zou2011unsupervised}
Will~Y Zou, Andrew~Y Ng, and Kai Yu.
\newblock Unsupervised learning of visual invariance with temporal coherence.
\newblock In {\em NIPS 2011 workshop on deep learning and unsupervised feature
  learning}, volume~3, 2011.

\end{thebibliography}
}

\end{document}